\documentclass[journal]{IEEEtran}
\ifCLASSINFOpdf
\else
\fi
\usepackage{amsmath}
\usepackage{amssymb}
\usepackage{algpseudocode}
\usepackage{algorithmicx}
\usepackage{algorithm}
\usepackage{makecell}
\usepackage{multicol}
\usepackage{multirow}
\usepackage{graphicx} % DO NOT CHANGE THIS：
\usepackage{subcaption}
\usepackage{tabularx,booktabs}
\newtheorem{theorem}{\textbf{Theorem}}
\newtheorem{proof}{\textbf{Proof}}

\begin{document}
\title{Deep Adversarial Inconsistent Cognitive Sampling for Multi-view Progressive Subspace Clustering}
%~\IEEEmembership{Member,~IEEE,}

\author{Renhao Sun}% <-this % stops a space
        
% <-this % stops a space
 % <-this % stops a space

\markboth{}%
{Shell \MakeLowercase{\textit{et al.}}: Bare Demo of IEEEtran.cls for IEEE Journals}



\maketitle

\begin{abstract}
Deep multi-view clustering methods have achieved remarkable performance. However, all of them failed to consider the difficulty labels (uncertainty of ground-truth for training samples) over multi-view samples, which may result into a non-ideal clustering network for getting stuck into poor local optima during training process; worse still, the difficulty labels from multi-view samples are always inconsistent, such fact makes it even more challenging to handle. In this paper, we propose a novel Deep Adversarial Inconsistent Cognitive Sampling (DAICS) method for multi-view progressive subspace clustering. A multi-view binary classification (easy or difficult) loss and a feature similarity loss are proposed to jointly learn a binary classifier and a deep consistent feature embedding network, throughout an adversarial minimax game over difficulty labels of multi-view consistent samples. We develop a multi-view cognitive sampling strategy to select the input samples from easy to difficult for multi-view clustering network training. However, the distributions of easy and difficult samples are mixed together, hence not trivial to achieve the goal. To resolve it, we define a sampling probability with theoretical guarantee. Based on that, a golden section mechanism is further designed to generate a sample set boundary to progressively select the samples with varied difficulty labels via a gate unit, which is utilized to jointly learn a multi-view common progressive subspace and clustering network for more efficient clustering. Experimental results on four real-world datasets demonstrate the superiority of DAICS over the state-of-the-art methods.
\end{abstract}


\IEEEpeerreviewmaketitle

\section{Introduction}

\IEEEPARstart{C}{lustering} is fundamental to computer vision and machine learning communities. With the emerging of big data, multi-view data clustering is vital to big data analytics, which aims to cluster the data into different groups by exploiting the complementary information from multiple feature spaces, with each of them corresponded to an individual view. To achieve nonlinear feature modeling, a number of Deep Neural Networks (DNN) based clustering methods \cite{zhou2018deep,yang2019deepspe,caron2018deep,jiang2018learn,shaham2018spectralnet,ji2017deep,zhang2019neural,zhang2019self} have achieved the desirable results in single-view scenario. To further enhance the performance, substantial
deep multi-view clustering methods \cite{andrew2013deep,huang2019multi,li2019reciprocal,hu2019deep,li2019deep,zhao2017multi,jiang2019dm2c,zhou2020end} are motivated.

Specifically, DCCA \cite{andrew2013deep} proposed a deep extension of the linear canonical correlation analysis (CCA) to learn deep feature  for multi-view clustering. MvDMF \cite{zhao2017multi} presented a deep multi-view clustering method through graph regularized semi-nonnegative matrix factorization, to capture the hidden information of each view while learning a common latent space to facilitate clustering.  MvSCN \cite{huang2019multi} performed deep multi-view spectral clustering, to project data objects from each view into a common space via deep neural network to preserve the local manifold invariance,  meanwhile achieving the consistency of pairwise view-specific representation via Siamese Network \cite{hadsell2006dimensionality}. Benefiting from adversarial learning \cite{goodfellow2014generative}, DAMC \cite{li2019deep}  pre-trained multi-view auto-encoder, and then jointly optimized auto-encoder and adversarial network to learn a common embedding for clustering. EAMC \cite{zhou2020end} jointly conducted  adversarial learning and attention mechanism, where they are leveraged to align the latent feature distributions and quantify the importance of each view.
\begin{figure}[t]
	\centering
	\includegraphics[width=0.75\columnwidth]{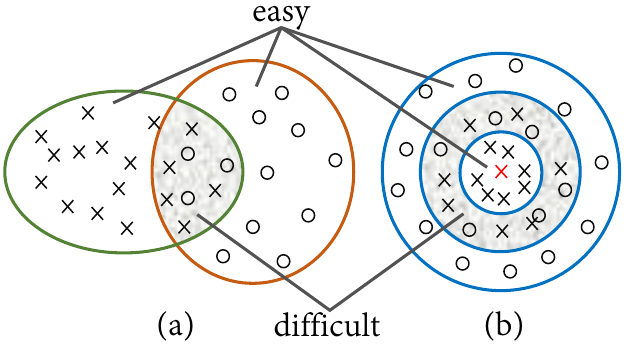} % Reduce the figure size so that it is slightly narrower than the column. Don't use precise values for figure width.This setup will avoid overfull boxes.
	\caption{(a) Samples \includegraphics[width=0.025\columnwidth]{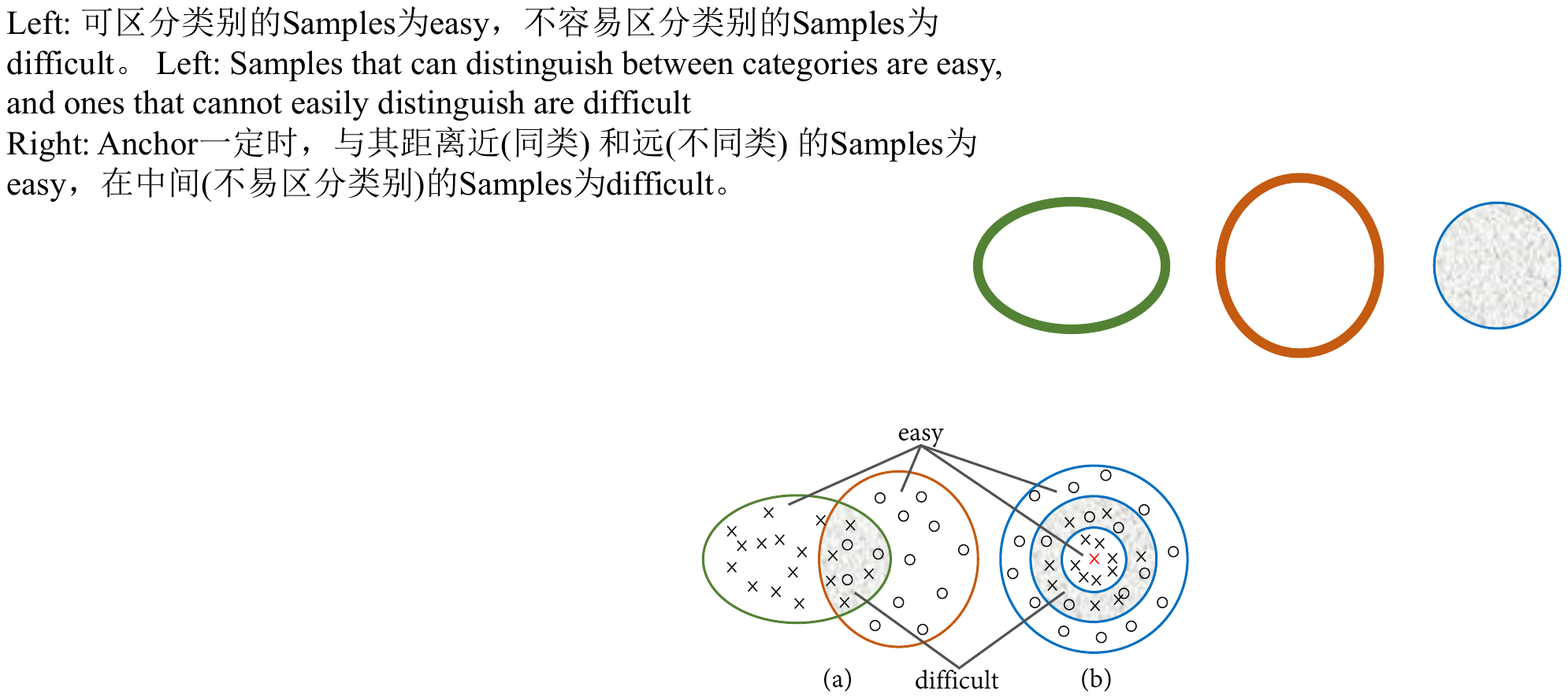} and \includegraphics[width=0.025\columnwidth]{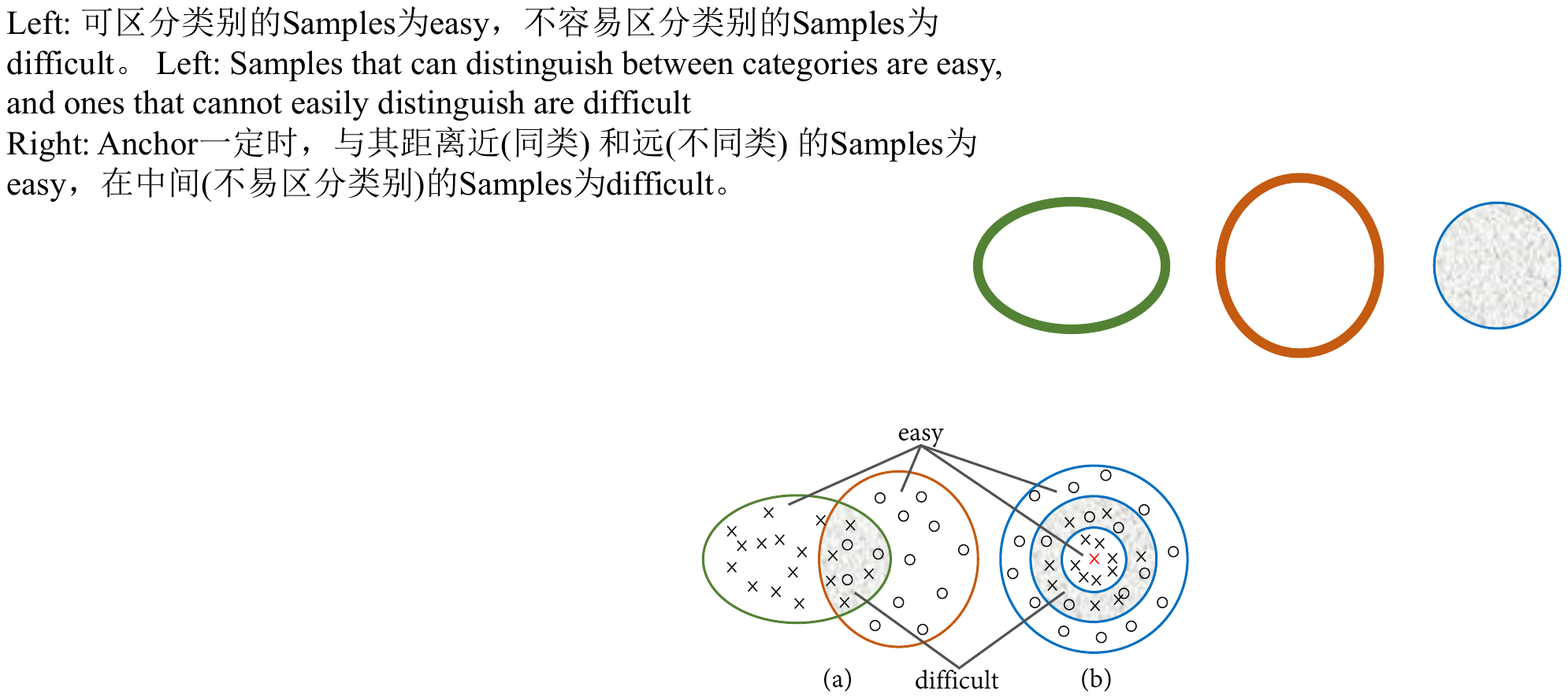} in the shadow area \includegraphics[width=0.025\columnwidth]{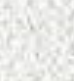} that cannot be easily distinguished between two classes (\includegraphics[width=0.03\columnwidth]{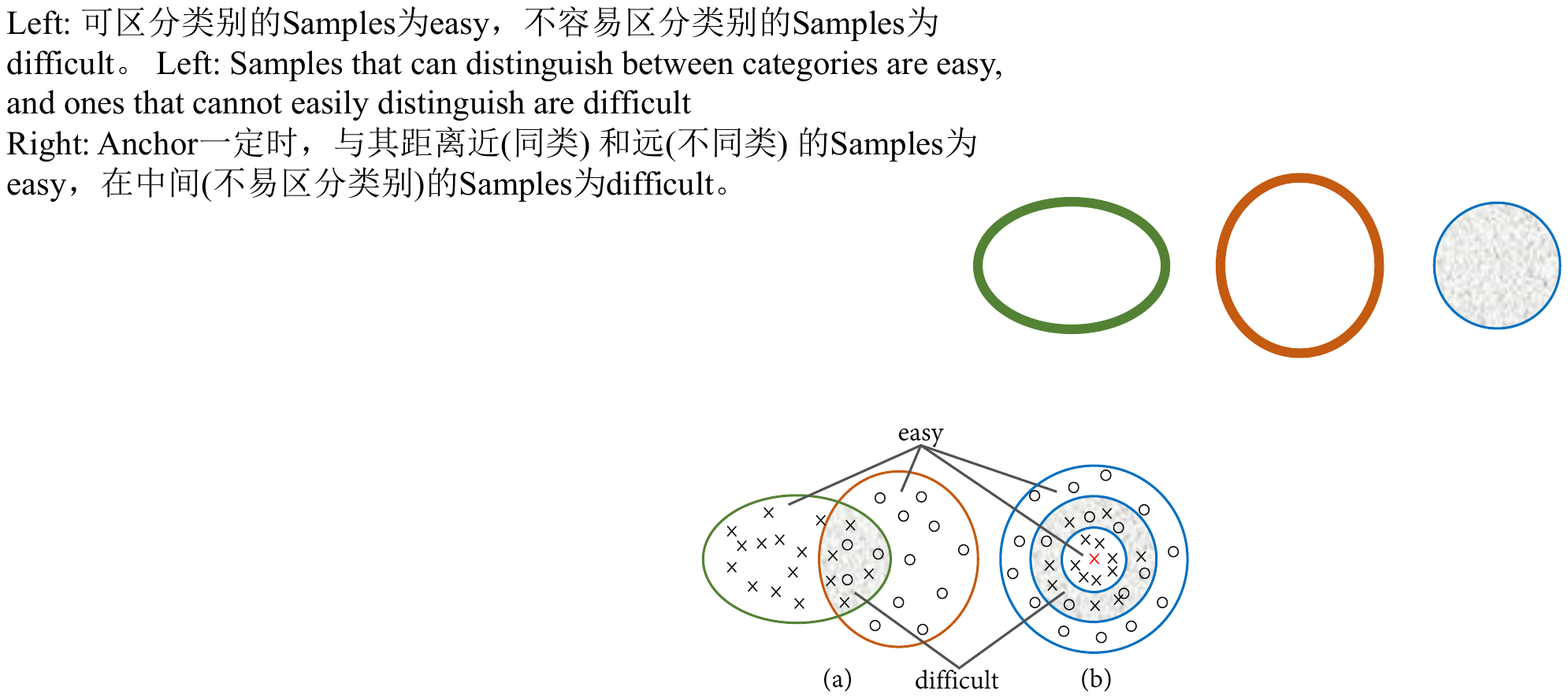} and  \includegraphics[width=0.025\columnwidth]{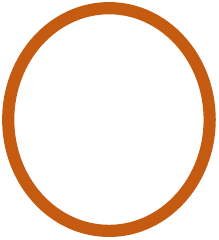} denote class 1 and class 2, respectively) are difficult, and ones in the other area that can be categorized are easy. (b) According to (a), we know that, once the anchor \includegraphics[width=0.025\columnwidth]{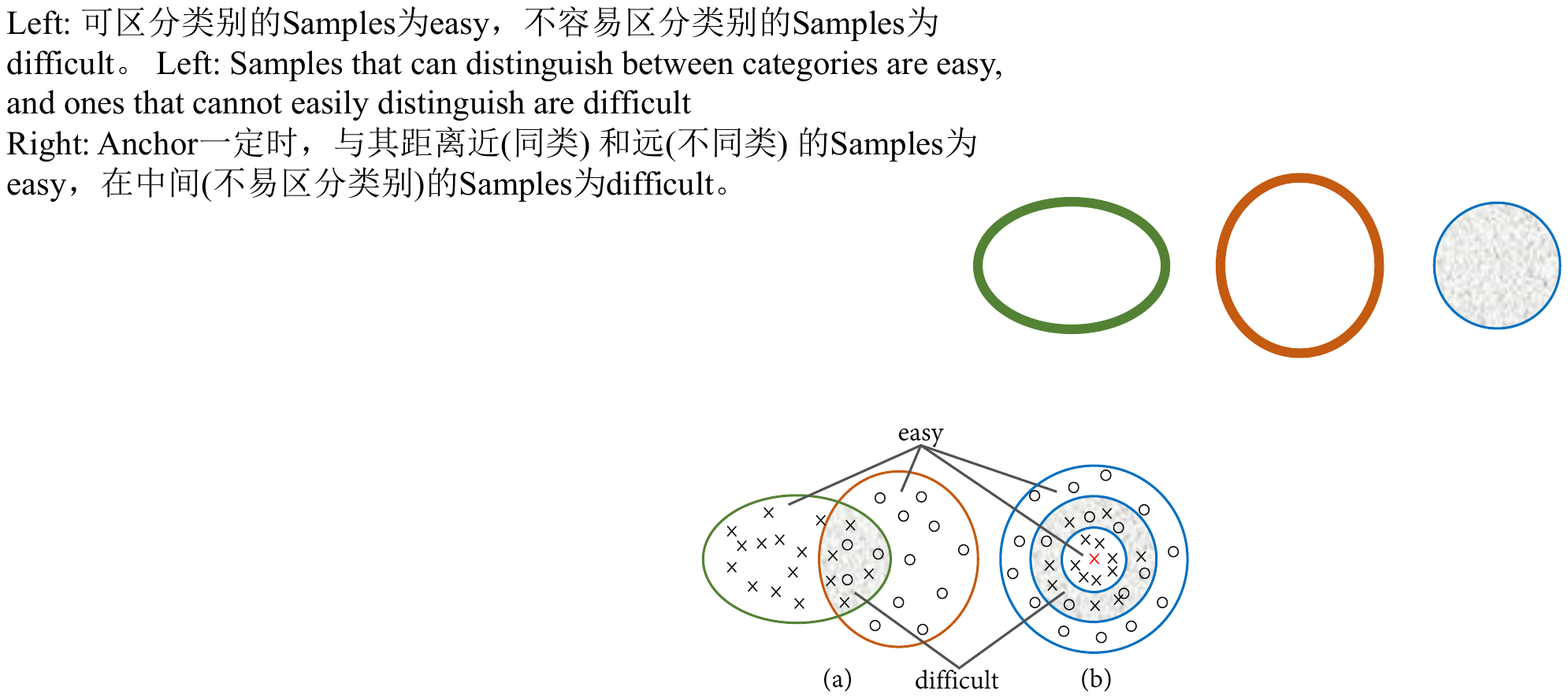} is fixed, samples that are close (belong to the same class) and far (belong to a different class) from it are easy, and those within its middle (belong to unknown classes) in the shadow area \includegraphics[width=0.025\columnwidth]{figure1-3.pdf} are difficult.}
	\label{fig1}
\end{figure}

\begin{figure*}[t]
	\centering
	\includegraphics[width=0.98\textwidth]{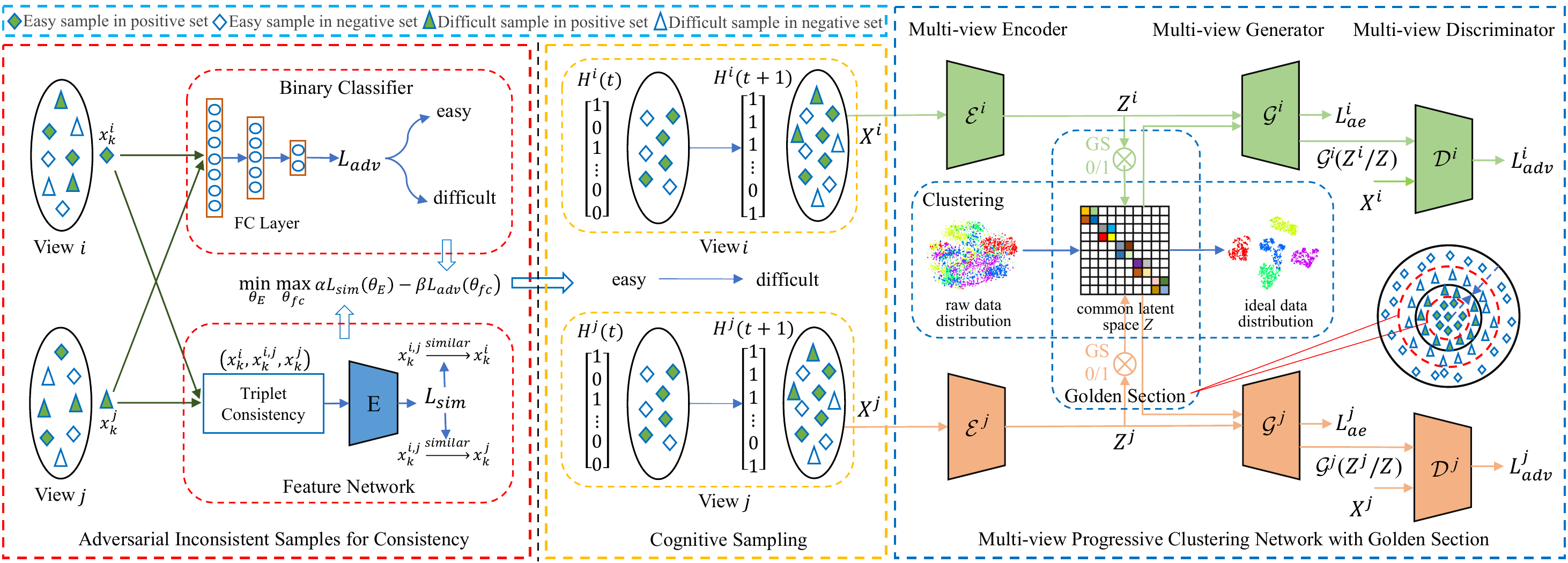} % Reduce the figure size so that it is slightly narrower than the column.
	\caption{The framework of DAICS (take the $i$-th and $j$-th views as example). DAICS consists of three components: the\textbf{ Adversarial Inconsistent Samples (AIS) module}, \textbf{Cognitive Sampling (CS) strategy} and Multi-view Progressive Clustering Network with \textbf{Golden Section (GS) mechanism}. The goal of the \textbf{AIS}  is to address the inconsistent difficulty labels of the multi-view samples via an adversarial minimax game of binary classifier and feature network. The \textbf{CS}  strategy is proposed to gradually train the multi-view clustering network from easy to difficult samples. The \textbf{GS} is developed with two states via a gate unit (0 or 1) to train multi-view clustering network including encoder, generator and discriminator, while learning a common progressive subspace. The details are shown in Fig. \ref{fig3}, for subspace clustering.}
	\label{fig2}
\end{figure*}

Despite the above progress, all of them ignore the difficulty labels (easy or difficult, see Fig. \ref{fig1}) of the training samples, \emph{i.e.,} uncertainty of the ground-truth for training samples, for clustering network training, which will affect the performance and generalization of the training network. It is apparent that the network will overfit for all easy training samples, and poorly trained with all training samples to be difficult. Numerous strategies have been proposed to address the above problem of the single-view clustering, such as self-paced learning \cite{ghasedi2019balanced} and cognitive learning \cite{jiang2018learn} strategies. However, it is not incrementally challenging for \emph{deep} multi-view clustering, due to \textbf{1)} the difficulty labels from multi-view samples are always inconsistent; \textbf{2)} how to collaborate within deep multi-view spaces for training sampling as per difficulty label is challenging and rarely solved.

To solve the above problem, we propose a novel Deep Adversarial Inconsistent Cognitive Sampling (DAICS) method for multi-view progressive subspace clustering (see Fig. \ref{fig2}), where the flowchart mainly comprises three components: the \textbf{Adversarial Inconsistent Samples (AIS)} module, the \textbf{Cognitive Sampling (CS)} strategy and the multi-view clustering network with the \textbf{Golden Section (GS)} mechanism. The basic idea of AIS is to convert samples with inconsistent difficulty labels into consistency by embedding samples from multiple views into a latent space via an adversarial minimax game between binary classifier and feature network. The binary classifier serves to \textit{distinguish} the difficulty labels of multi-view inconsistent samples to train the deep consistent feature embedding network, and the feature network learns a common feature representation from the multi-view inconsistent samples to \textit{confuse} the binary classifier. The multi-view CS strategy is developed to select the samples from easy to difficult according to the probability of the samples for the clustering network training. As indicated by \cite{ghasedi2019balanced}, selecting from the easy-to-difficult sampling strategy could avoid getting stuck in poor local minima while capturing the intrinsic patterns of multi-view samples. For the network efficiency, the GS mechanism is designed with two states via a gate unit. One is to learn the latent representation and the clustering network of each view relying on easy samples. The other learns a multi-view common progressive subspace to coordinate all views with both easy and difficult samples, where the samples are progressively processed, leading to a multi-view progressive subspace clustering.

The major technical contributions are summarized below:
\begin{quote}	
	\begin{itemize}
		\item A novel DAICS method is proposed to consider the difficulty labels of the samples for deep multi-view subspace clustering.
		\item A  multi-view cognitive sampling strategy is developed to select samples from easy to difficult with getting stuck in poor local minima being avoided.
		\item To achieve the clustering network training efficiency, we present a golden section mechanism to learn a multi-view common progressive subspace via a gate unit (0 or 1).
	\end{itemize}
\end{quote}
The extensive ablation studies validate the advantages of DAICS.

\section{Related Works}
\subsection{Subspace clustering}
To date, massive subspace clustering algorithms \cite{li2018geometric,yamaguchi2019subspace,li2017structured,elhamifar2013sparse,vidal2011subspace,wang2015robust,wang2018multiview,wang2016iterative} have been proposed, among them, spectral subspace clustering is one of the popular linear algorithms to cluster high-dimensional data \cite{vidal2011subspace}. The most challenging problem in spectral subspace clustering is the construction of affinity matrix. Existing studies on affinity matrix in spectral subspace clustering could be classified into three main groups: 1) matrix factorization based algorithms \cite{wang2018multiview,wang2016iterative}, 2) model based algorithms \cite{yamaguchi2019subspace}, 3) self-expression based algorithms \cite{liu2012robust,vidal2014low}. These algorithms perform spectral subspace clustering on the affinity matrix to cluster the high-dimensional data points. Nevertheless, the processing of high-dimensional data always takes expensive time and computation. The sparse subspace clustering algorithms \cite{li2018geometric,li2017structured,elhamifar2013sparse} have been developed to infer the clustering of data points into a low-dimensional subspace by solving a sparse optimization program whose solution is used in spectral subspace clustering algorithms. However, these algorithms are not able to model the non-linear and high-dimensional complex real-world data because of they can only cluster linear subspace. To resolve it, the deep subspace clustering algorithms have been proposed and attracted plenty of attention, which will be detailed in the next section.

\subsection{Deep Clustering}
Inspired by deep learning \cite{wu2018cycle,wang2021survey},  the deep clustering algorithms have been developed to model the non-linear and high-dimensional complex data, and have enough capacity to deal with the large-scale datasets. There are two main branches of the existing methods. One branch is based on the auto-encoder networks, which is aiming to learn a common latent representation by reconstructing the input samples for a better clustering. Ji \textit{et al. } \cite{ji2017deep} proposed an unsupervised subspace clustering method that was built upon deep auto-encoders, and introduced a self-expressive layer to learn pairwise affinities for clustering. Based on the above, Zhang \textit{et al.} \cite{zhang2019neural} presented a neural collaborative subspace clustering algorithm to construct negative and positive two confidence affinity matrices, which could supervise each other to promote training. Yang \textit{et al.} \cite{yang2019deepspe} and Zhang \textit{et al.} \cite{zhang2019self} developed a self-supervised module that exploited the output of spectral clustering \cite{ng2002spectral} to achieve optimal clustering results. For the model robustness, Jiang \textit{et al.} \cite{jiang2019duet} contributed a duet robust deep subspace clustering, to handle contaminated data and enhance the robustness from both the self-expressive and the data reconstruction perspective with two regularization norms. In addition, a robust deep subspace clustering framework \cite{jiang2018learn} was proposed, based on the principle of human cognitive process, learning gradually samples from easy to difficult and less to more.

The other is based on the joint deep learning framework of the auto-encoders and the Generative Adversarial Networks (GANs) \cite{goodfellow2014generative}, compared to the previous branch, a discriminator network is introduced to further capture the data distribution and disentangle the common latent representation for a better clustering. In the single-view clustering methods, a ClusterGAN \cite{mukherjee2019clustergan} is proposed as a new mechanism using GANs by sampling latent variables for clustering, to preserve latent representation interpolation across categories for clustering. Almost contemporaneously, due to deep models with large number of parameters prone to overfitting, Dizaji \textit{et al.} \cite{ghasedi2019balanced} presented a deep generative adversarial clustering network with a balanced self-paced learning algorithm to tackle the problem. In the multi-view clustering methods, Li \textit{et al.} \cite{li2019deep} developed a deep adversarial multi-view clustering (DAMC) network to learn the intrinsic structure of multi-view samples, which consists of a deep auto-encoder and an adversarial learning process for each view, for a better clustering. However, the more complex the models are, the larger the number of parameters are. The deep models prone to the overfitting leading to get stuck in poor local minima for multi-view clustering. Therefore, starting from the difficulty labels of the multi-view samples, we propose a deep adversarial inconsistent cognitive sampling (DAICS) method to efficiently train deeper networks for multi-view progressive subspace clustering.

\subsection{Self-paced Learning Algorithms}
Self-paced learning derives from curriculum learning \cite{bengio2009curriculum}, a human-like learning principle where the easier instances are learned first, and then  more difficult instances are gradually introduced to the learning process. Kumar \textit{et al.} \cite{kumar2010self} proposed a self-paced learning algorithm to learn a new parameter vector from easy samples to difficult samples, compared to curriculum learning, the self-paced learning automatically adjusted the sample difficulty. Jiang \textit{et al.} \cite{jiang2014self} developed a extension of the self-paced learning algorithm, the diversity is introduced to a general regularization term. Many works \cite{li2016multi, liang2016self, zhang2015self} further considered self-paced learning into various tasks to avoid getting stuck in poor local minima and improve the generalization for models. For the clustering tasks, the self-paced learning was used to the single-view clustering to improve the robustness of the models. Unlike the single-view clustering, multi-view clustering relies on the complementary information from the samples of multiple views. The application of cognitive learning on the multi-view clustering is not a simple extension from one on the single-view clustering. The process of which is demanded to collaborate with the difficulty labels (easy or difficult) of multi-view samples.

\section{The Proposed Method}
In this section, our proposed DAICS method is illustrated in Fig. \ref{fig2}, which clusters a set of $n$ data objects with $V$ views $D = \{\textbf{X}^1,...,\textbf{X}^i,...,\textbf{X}^V \}$ into $c$ clusters, where $\textbf{X}^i \in \mathbb{R}^{d_i \times n} $ denotes the samples of the dimension $d_i$ from the $i$-th view. The DAICS that consists of three components: the adversarial inconsistent samples (AIS) module, the cognitive sampling (CS) strategy and the multi-view clustering network with the golden section (GS) mechanism will be elaborated in detailed, together with their implementation.

\subsection{Adversarial Inconsistent Samples}
Given $\textbf{X}^{i}=\{x_1^i,...,x_n^i\}$ of the $i$-th view. Following \cite{gao2018self},  as shown in Fig. \ref{fig3}(a),  we obtain the positive sample set $\textbf{P}^i$, w.r.t. $\textbf{R}_{\mathrm{P}^{i}}$ and the negative sample set $\textbf{N}^i$ w.r.t. $\textbf{R}_{\mathrm{N}^{i}}$ by K Nearest Neighbors (K-NN) algorithm. For $\textbf{P}^i$ and $\textbf{N}^i$, we further determine the difficulty label of $x_k^i$ by the distance from a randomly selected anchor object such as $x_{a}^{i}$, resulting into the distributions of samples from easy to difficult for $\textbf{R}_{\mathrm{P}^{i}}$ and $\textbf{R}_{\mathrm{N}^{i}}$.  For $\textbf{P}^{i}$, if the distance to $x_{a}^{i}$ is small, it is more likely to be easy, \emph{i.e.,} certain to be the same group as $x_{a}^i$, while for $\textbf{N}^i$, the large distance to $x_{a}^{i}$ is more likely to be easy, \emph{i.e.,} certain to be the different group from $x_{a}^i$. To this end,  the difficulty label of $x_{k}^i$ is defined as
\begin{equation}\label{eq1}
	y_{k}^{i}=\left\{\begin{matrix}
		0,   &  d_{k}^{\mathrm{N}^i} > \mu d_{max}^{\mathrm{N}^i} \ \ or \ \ d_{k}^{\mathrm{P}^i} < \mu d_{max}^{\mathrm{P}^i}\\
		\\
		1,   &  otherwise
	\end{matrix}\right.,
\end{equation}
where $y_{k}^{i}= 0$ $(1)$ indicates that the label of sample $x_k^i$ is easy (difficult). $d_{k}^{\mathrm{N}^i}$ and $d_{k}^{\mathrm{P}^i}$ denote the distance between $x_k^i$ and $x_{a}^{i}$ in the set $\textbf{N}^i$ and $\textbf{P}^i$, respectively. $d_{max}^{\mathrm{N}^i}$ and $d_{max}^{\mathrm{P}^i}$ denote the maximum distance from the anchor $x_{a}^{i}$ in $\textbf{N}^i$ and $\textbf{P}^i$, respectively. $\mu$ is a boundary factor for easy and difficult samples.

As it is widely known, the difficulty labels of multi-view samples are always inconsistent. We develop an adversarial strategy to make the difficulty labels of such pairs of samples to be consistent. Our model comprises a binary classifier that generates the difficulty labels of the fused sample representations throughout a fully connected network, equipped with an adversarial cross-entropy loss to yield a pseudo difficulty label. Another feature embedding network is learned to achieve a common representation for inconsistent samples. The difficulty labels are obtained via a hinge loss. Afterwards, such two labels are processed throughout an adversarial minimax game to yield the consistent difficulty labels.
In what follows, we discuss the details for the binary classifier and the feature embedding network.
\subsubsection{Binary Classifier}
After determining the difficulty labels of all samples via Eq. (\ref{eq1}), it is trivial to collect the samples with inconsistent labels. Assume any pair of inconsistent samples $(x_k^i,x_k^j)$, such that $y_k^i\not=y_k^j$.
%We define a binary classifier to coordinate the difficulty degree, which is regarded as "discriminator" in generative adversarial network(GAN).

We propose a binary classifier to minimize the binary classification throughout an adversarial loss, as formulated below:
\begin{equation}\label{eq2}
	\mathcal{L}_{adv}(\theta_{fc})=-\frac{1}{K}\sum_{y_k^i\not=y_k^j}\ell({\rm log}f(x_k^i)+{\rm log}(1-f(x_k^j))),
\end{equation}
where $\mathcal{L}_{adv}$ denotes the cross-entropy loss to classify all $K$ pairs of samples with inconsistent difficulty labels into the binary value (0/1) for easy or difficult. $\ell$ is a factor of the pseudo difficulty label, such that $0 \leq\ell\leq 1$.  $f(\cdot)$ is a binary classifier including three fully connected layers with the parameters $\theta_{fc}$, with its output to be 0 or 1, which is generated to guide the training process of the deep consistent embedding feature network that is discussed in the next.

\subsubsection{Feature Network}
Based on the pair $\{x_k^i,x_k^j\}$, we construct a triplet $\{x_k^i,x_k^{i,j},x_k^j\}$, as the dimension for all three samples are distinct, we expect that $x_k^{i,j}$ is similar to both $x_k^i$ and $x_k^j$ within the output space of the feature network. Motivated by this, we define a deep feature embedding network \textbf{E}, formulated as
\begin{equation}\label{eq3}
	\Delta x_k^{i,j} = \arg \min_{x_k^{i,j}}\left \| F(x_k^{i,j})-F(x_k^j)\right \|_{2}^{2},
\end{equation}
\textit{s.t.}
\begin{center}
	$\left \| F(x_k^{i,j}) - F(x_k^i)\right \|_{\infty }< \xi,$
\end{center}
where $F(\cdot)$ denotes the output for $\textbf{E}$ parameterized as $\theta_{E}$. $\xi$ denotes a factor. According to Eq. (\ref{eq3}), the following constraint is satisfied:
\begin{equation}\label{eq4}
	\left \| F(x_k^{i,j})-F(x_k^i)\right \|_{2}^{2}>\left \| F(x_k^{i,j})-F(x_k^j)\right \|_{2}^{2}.
\end{equation}
However, the common feature representation extracted from the sample $x_k^{i,j}$ is not only similar to that of $x_k^i$ but also similar to that of $x_k^j$. Hence, we aim at the following:
\begin{equation}\label{eq5}
	\left \| F(x_k^{i,j})-F(x_k^i)\right \|_{2}^{2}<\left \| F(x_k^{i,j})-F(x_k^j)\right \|_{2}^{2}.
\end{equation}
Eq. (\ref{eq4}) and Eq. (\ref{eq5}) all together comprise an adversarial learning process, to facilitate mutual learning over easy and difficult samples. Based on that,  the feature similarity loss is defined as follows:
\begin{equation}\label{eq6}
	\begin{aligned}
		\mathcal{L}_{sim}(\theta_{E}) = \frac{1}{K}\sum_{y_k^i\not=y_k^j}{\rm max}(0,m+\left \| F(x_k^{i,j})-F(x_k^i)\right \|_{2}^{2}
		\\-\left \| F(x_k^{i,j})-F(x_k^j)\right \|_{2}^{2}),
	\end{aligned}
\end{equation}
where $\mathcal{L}_{sim}$ denotes a hinge loss \cite{chen2017beyond,zhong2019adversarial}. $m$ is the margin, indicating a similarity preference factor.
When $m$ is a smaller value, the network can obtain the better performance.

\subsubsection{Adversarial Learning} The final consistent difficulty labels for $(x_{k}^{i}, x_{k}^{j})$ are obtained via the adversarial learning process of the binary classification loss $\mathcal{L}_{adv}$ and the feature similarity loss $\mathcal{L}_{sim}$, to jointly learn the binary classifier $f(\cdot)$ and the feature network $F(\cdot)$, formulated as \textbf{a minimax game} with the following objectives:

\begin{equation}\label{eq7}
	\hat{\theta}_{E}=\arg \min_{\theta_{E}}(\alpha \mathcal{L}_{sim}(\theta_{E})-\beta \mathcal{L}_{adv}(\hat{\theta}_{fc})),
\end{equation}

\begin{equation}\label{eq8}
	\hat{\theta}_{fc}=\arg \max_{\theta_{fc}}(\alpha \mathcal{L}_{sim}(\hat{\theta}_{E})-\beta \mathcal{L}_{adv}(\theta_{fc})).
\end{equation}
We summarize the above process in Algorithm \ref{al1}.

\begin{algorithm}[t]
	\caption{Pseudocode of optimizing our AIS}
	\hspace*{0.02in} {\bf Initialization:} $k$ samples of the i-th view: $\{x_1^i,...,x_k^i\}$;
	
	\quad \ \ $k$ samples of the j-th view: $\{x_1^j,...,x_k^j\}$;
	
	\quad \ \ Hyperparameters: $t, m, \ell, \alpha, \beta$;
	
	\quad \ \ Batch size: $b$;
	
	\hspace*{0.02in} {\bf Update until convergence:}
	\begin{algorithmic}[1]
		\For{$t$ steps}
		\State Update the parameters $\theta_{E}$ of the embedding feature
		\Statex \quad \ \ network by descending their stochastic gradients:
		\State \quad \quad $\theta_{E}\gets\theta_{E}-\eta\cdot\nabla_{\theta_{E}}\frac{1}{b}(\alpha \mathcal{L}_{sim}-\beta \mathcal{L}_{adv})$
		\EndFor
		\State Update the parameters $\theta_{E}$ of the binary classifier by ascending their stochastic gradients:
		\State \quad \quad \quad \ \ $\theta_{fc}\gets\theta_{fc}+\eta\cdot\nabla_{\theta_{fc}}\frac{1}{b}(\alpha \mathcal{L}_{sim}-\beta \mathcal{L}_{adv})$
		\State \Return Learned embedding feature $F(\cdot)$ and binary classifier $f(\cdot)$
	\end{algorithmic}
	\label{al1}
\end{algorithm}

\subsection{Cognitive Sampling}
\begin{figure}[t]
	\centering
	\includegraphics[width=0.9\columnwidth]{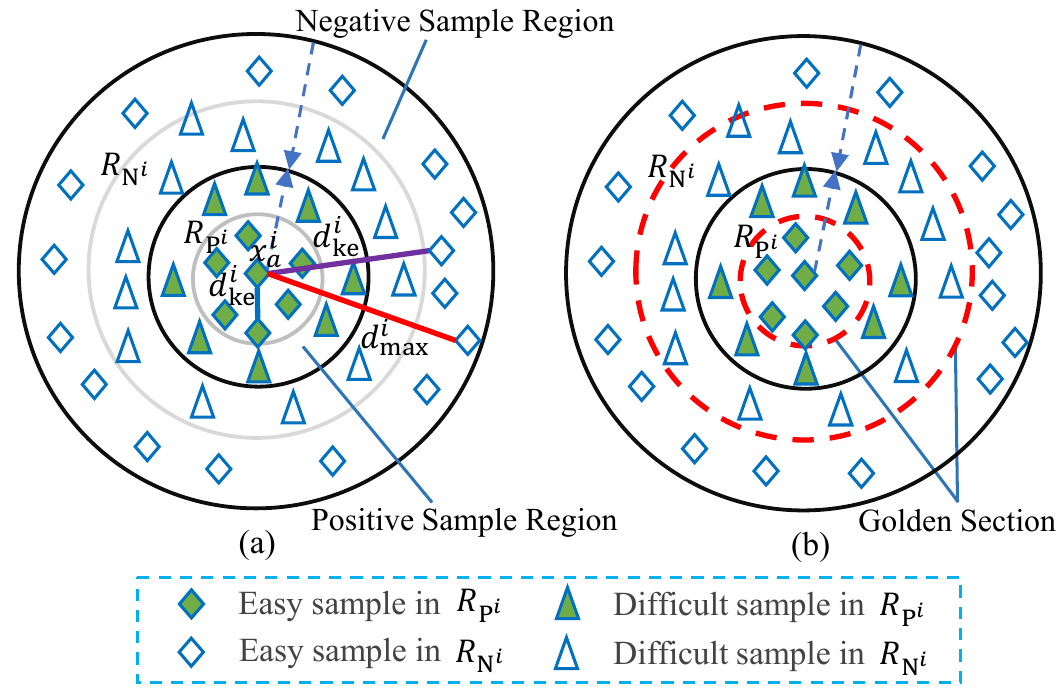} % Reduce the figure size so that it is slightly narrower than the column. Don't use precise values for figure width.This setup will avoid overfull boxes.
	\caption{For the $i$-th view, (a): $R_{\mathrm{P}^i}$ and $R_{\mathrm{N}^i}$ denote the positive and negative sample region. The difficulty labels of multi-view samples are defined via Eq. (\ref{eq1}) and the samples from easy to difficult are gradually selected via Eqs. (\ref{eq9}), (\ref{eq10}) and (\ref{eq11}). (b): The red slash circles in $R_{\mathrm{P}^i}$ and $R_{\mathrm{N}^i}$ denote the golden section to divide multi-view clustering network to two states via a gate unit to learn a common progressive subspace (see Eq. (\ref{eq12}) for the details).}
	\label{fig3}
\end{figure}

Upon the consistent difficulty labels of the training multi-view samples, we aim at selecting from easy to difficult, to avoid trapping in poor local minima resulting into undesirable generalization. However, as illustrated in Fig. \ref{fig3}(a), the distributions of easy and difficult samples are mixed together, and hence not trivial to solve. To this end, we define a sampling probability, such that the probability of easy sample is larger than difficult one.

For both easy and difficult samples, we consider them in both positive and negative sets. Specifically, if the label of the sample $x_k^i$ is easy, the probability $p_{ke}^i$ is defined as follows:
\begin{equation}\label{eq9}
	p_{ke}^{i}=\left\{\begin{matrix}
		\frac{d_{ke}^{i}}{d_{\max }^{i}},   &  x_{k}^{i} \in \textbf{N}^{i} \\
		\\
		1-\frac{d_{ke}^{i}}{d_{\max }^{i}},  & x_{k}^{i} \in \textbf{P}^{i}
	\end{matrix}\right.,
\end{equation}
where $d_{ke}^{i}$ denotes the distance between $x_k^i$ and the anchor $x_a^i$. $d_{\max }^{i}$ denotes the maximum distance between $x_k^i \in \textbf{N}^i$ and $x_a^i$.  According to Eq. (\ref{eq9}), for $x_k^i \in \textbf{N}^i$, the larger distance to $x_a^i$, the larger probability to be sampled (more easily determined to be a different group as $x_a^i$). For $x_k^i \in \textbf{P}^i$, the smaller distance to $x_a^i$, the larger probability to be sampled (more easily determined to be the same group as $x_a^i$).

If the label of the sample $x_k^i$ is difficult, we define the following:
\begin{equation}\label{eq10}
	p_{kd}^{i}=\frac{\left|d_{kd}^{i}-d_{med}^{i}\right|}{\sum_{l=1}^{n_{d}} d_{l}^{i}},
\end{equation}
where $d_{kd}^{i}$ denotes the distance between $x_k^i$ and $x_a^i$. $d_{med}^{i}$ denotes the median distance among the distances of all $n_d$ difficult samples to $x_a^i$.  It is easily seen from Eq. (\ref{eq10}) that the smaller distance between $x_k^i$ and easy samples in both positive and negative sets, the larger $p_{kd}^i$ is, otherwise, the smaller it is. The last question is whether $p_{ke}^i > p_{kd}^i$ to ensure the sequence from easy to difficult, which is answered in the following theorem.
\begin{theorem}
	$p_{ke}^i > p_{kd}^i$
\end{theorem}
\begin{proof}
	Empirically,  $n_d \gg 1$ to ensure ${\sum_{l=1}^{n_{d}} d_{l}^{i}} > d_{\max }^{i}$, hence we consider the numerator of Eqs. (\ref{eq9}) and (\ref{eq10}). If $x_k^i$ is easy and $x_k^i \in \textup{\textbf{N}}^i$, as shown in Fig. \ref{fig3}(a), it is obvious that, for any difficult sample, we have $0< d_{kd}^i < d_{ke}^i$,  $0< d_{med}^i < d_{ke}^i$, hence it yields $\left|d_{kd}^{i}-d_{med}^{i}\right| < d_{ke}^i$.
	If $x_k^i \in \textup{\textbf{P}}^i$, the numerator of $p_{ke}^i$ is $d_{max}^i-d_{ke}^i$, it is apparent that $0< d_{kd}^i < d_{max}^i$, $0< d_{med}^i < d_{max}^i$. Meanwhile
	$0< d_{ke}^i < d_{kd}^i$, $0< d_{ke}^i < d_{med}^i$. Hence, we have $\left|d_{kd}^{i}-d_{med}^{i}\right| < d_{max}^i-d_{ke}^i$.
	It finally leads to $p_{kd}^i < p_{ke}^i$.  $\hfill\blacksquare$
\end{proof}

For $V$ views, we define $\textbf{H}^i(t)=\{h_1^i(t),...,h_n^i(t)\}$  with $h_k^i(t)$ as a binary-value $(0/1)$ to decide whether $x_k^i$ is selected.
To ensure the selection the samples from easy to difficult among $V$ views, $h_k^i(t)$ is defined by
\begin{equation}\label{eq11}
	h^{i}_k(t)=\left\{\begin{matrix}
		1,  & \lambda(t) \leq \frac{1}{V}\sum_{i=1}^{V} p_{k}^{i} \\
		\\
		0, & otherwise %\lambda(t) > \frac{1}{V}\sum_{i=1}^{V} p_{k}^{i}
	\end{matrix}\right.,
\end{equation}
where $\lambda(t) \in [0,1)$ is a self-adjusting sampling pace. $p_k^i$ is the sampling probability of $x_k^i$ as per Eqs. (\ref{eq9}) and (\ref{eq10}).
For each view, \textit{e.g., }the $i$-th view, we average the sampling probability for all $V$ views.
When the training process started, $\lambda(t)$ is a higher value so that only easy samples are selected. With the  iteration number $t$ increases, $\lambda(t)$ is decreasing to select difficult samples. Specifically, the difficult samples closer to easy samples are selected first, followed by others. Next, we discuss how to train the multi-view clustering network and learn a common progressive subspace via the easy and difficult samples.

\subsection{Multi-view Progressive Clustering Network with Golden Section}
Our multi-view clustering network adopts the architecture of auto-encoder combined with GAN \cite{xu2019adversarial,li2019deep}, which consists of multi-view encoder network, multi-view generator network and multi-view discriminator network for each view.

\textbf{Multi-view encoder network $\mathcal{E}^{i}$:} Our multi-view encoder networks are different for the diverse feature dimensions of each view, there are multi-layer convolution neural networks with distinctive parameters. For the $i$-th view, given $\textbf{X}^i=\{x_1^i,...,x_n^i\}$ , the multi-view encoder $\mathcal{E}^{i}$ aims to learn the latent representations $\textbf{Z}^i=\{z_1^i,...,z_n^i\}$. Specifically, it maps the $d_i$-dimensional input sample $x^i$ to a low-dimensional latent representation $z^i$. This mapping could be represented as $\textbf{Z}^i=E_i(\textbf{X}^i;\Theta_{\mathcal{E}^i})$, where $E_i(\cdot)$ refers to the $i$-th view’s encoder network parameterized by $\Theta_{\mathcal{E}^i}$.

\textbf{Multi-view generator network $\mathcal{G}^{i}$:} Our multi-view generator network $\mathcal{G}^{i}$ is set as a symmetrical architecture of multi-view encoder network $\mathcal{E}^{i}$ for the $i$-th view, which consists of opposite multi-layer convolution neural networks with distinctive parameters. Specifically, the network  $\mathcal{G}^{i}$ can generate the reconstructed samples $\hat{\textbf{X}}^i$ with the latent representations $\textbf{Z}^i=\{z_1^i,...,z_n^i\}$ corresponding to the $i$-th view. We suppose $\hat{\textbf{X}}^i=G_i(\textbf{Z}^i;\Theta_{\mathcal{G}^i})$ , where $G_i(\cdot)$ refers to the $i$-th view’s generator network parameterized by $\Theta_{\mathcal{G}^i}$.

\textbf{Multi-view discriminator network  $\mathcal{D}^{i}$:} Our multi-view discriminator network
$\mathcal{D}^{i}$ consists of 3 fully connected layers, which can distinguish a generated sample or a real sample. $D^{i}(\cdot)$ parameterized by $\Theta_{\mathcal{D}^i}$ feeds back the result to generator network and updates the parameters of generator. By this means, the discriminator $\mathcal{D}^{i}$ works as a regularizer to guide the training of our multi-view encoder network $\mathcal{E}^{i}$, which enhances the robustness of embedding representations and avoids the overfitting issue effectively.

Our goal is to learn a multi-view common subspace for clustering, while conducting multi-view clustering network collaboration throughout the common subspace. One natural question is about the training set for the multi-view common subspace learning and the clustering network learning. First, the overfitting will occur for the common subspace learning with easy training samples only, hence we proposed to learn that once handling difficult training samples. Second, to avoid trapping into poor local optima for clustering network, the easy samples are ideal for initial training, then the difficult ones are gradually joining into the training process for an effective multi-view progressive clustering network.

\subsubsection{Golden Section}
We design a golden section (GS) mechanism, which detects whether the number of the training samples reach the golden section via a gate unit, seen as the two red slash circles in Fig. \ref{fig3}(b). According to Eq. (\ref{eq11}), for the $t$-th iteration of the training process, the number of the input samples is calculated as $\sum_{k=1}^{n} h_{k}^{i}(t)$. Therefore, we have:
\begin{equation}\label{eq12}
	GS =\left\{\begin{matrix}
		1, & \sum_{k=1}^{n} h_{k}^{i}(t) > \sigma N^i\\
		\\
		0, &  \textit{otherwise}
	\end{matrix}\right.,
\end{equation}
where $N^i$ denotes the number of all samples for the $i$-th view. $\sigma$ is a golden section factor, which is exactly the boundary factor $\mu$ in Eq. (\ref{eq1}), the value is $(\sqrt{5}-1)/2\approx 0.618$. Two states are yielded via a gate unit. $GS=0$ (closed state) indicates that the number of the samples is below the golden section. The network only learns the corresponding latent representations of each view; $GS=1$ (open state) indicates exceeding the golden section, where the multi-view common progressive subspace is learned upon multi-view latent representations and the different network modules from each view.
The above two states result into the different loss functions for the multi-view clustering network to be optimized. We discuss each of them in the next.

\subsubsection{The Auto-Encoder Loss}
The auto-encoder loss is measured by $L_2$ distance between the reconstructed sample and the real sample, when $GS=0$, the loss for the $i$-th view is
\begin{equation}\label{eq13}
	\hat{\mathcal{L}}_{ae}^{i}(\Theta_{\mathcal{E}^i};\Theta_{\mathcal{G}^i}) = \left \| \textbf{X}^i-\hat{\textbf{X}}^i\right \|^{2}.
\end{equation}
For $GS = 1$, the common subspace $\textbf{Z}$ depends on the latent representations of all views, meanwhile learned throughout the network training. Then, we have
\begin{equation}\label{eq14}
	\tilde{\mathcal{L}}_{ae}^{i}(\Theta_{\mathcal{E}^i};\Theta_{\mathcal{G}^i}) = \left \| \textbf{X}^i-\hat{\textbf{X}}^i\right \|^{2}+\lambda_i (\left \| \textbf{X}^i-\tilde{\textbf{X}}^i\right \|^2 + \left \| \textbf{Z}^i-\textbf{Z}\right \|^2),
\end{equation}
where $\lambda_i$ denotes a factor with the value $1/V$. $\hat{\textbf{X}}^i=G_i(\textbf{Z}^i;\Theta_{\mathcal{G}^i})$ and  $\textbf{Z}^i=E_i(\textbf{X}^i; \Theta_{\mathcal{E}^i})$. The reconstructed sample $\tilde{\textbf{X}}^i = G_i(\textbf{Z};\Theta_{\mathcal{G}^i})$. We learn the common subspace $\textbf{Z}$ by minimizing the reconstruction loss with $\textbf{Z}^i$, while confusing discriminator between $\textbf{X}^i$ and $\hat{\textbf{X}}^i$ when $GS = 0$. Besides, $\textbf{X}^i$ and $\tilde{\textbf{X}}^i$ is added when $GS = 1$, modeled by the following adversarial loss.

\begin{algorithm}[t]
	\caption{The proposed DAICS algorithm}
	\hspace*{0.02in} {\bf Initialization:} The dataset D=$\{\textbf{X}^1,...,\textbf{X}^i,...,\textbf{X}^V\}$;
	
	\quad \ \ Hyperparameters: $t, m, \ell, \alpha, \beta$;
	
	\hspace*{0.02in} {\bf Update until convergence:}
	\begin{algorithmic}[1]
		\State Adopt the AIS algorithm (Algorithm 1) to obtain the samples with difficult consistency
		\For{$t$ steps}
		\State Use the CS algorithm to gradually increase the sam-
		\Statex \quad \ \ ple from easy to difficult
		\If {$GS$ == 0}
		\State Update $\Theta_{\mathcal{E}^i}$ and $\Theta_{\mathcal{G}^i}$ via the Eq.\ref{eq13}:
		\Statex \quad \quad \quad \quad \quad $\Theta_{\mathcal{E}^i}, \Theta_{\mathcal{G}^i}\gets\min\hat{\mathcal{L}}_{ae}^{i}(\Theta_{\mathcal{E}^i};\Theta_{\mathcal{G}^i})$
		\State Update $\Theta_{\mathcal{D}^i}$ via the Eq.\ref{eq15}: 
		\Statex \quad \quad \quad \quad \quad $\Theta_{\mathcal{D}^i}\gets\max\mathbb{E}_{x^i\sim P(\textbf{X}^i)}[{\rm log}D_i(x^i)]$
		\Statex \quad \quad \quad \quad \quad \quad \quad \quad \quad \quad$+\mathbb{E}_{\hat{x}^i\sim P(\hat{\textbf{X}}^i)}[1-{\rm log}D_i(\hat{x}^i)]$
		\State Update $\Theta_{\mathcal{E}^i},\Theta_{\mathcal{G}^i}$ via the Eq.\ref{eq15}:
		\Statex \quad \quad \quad \quad \quad $\Theta_{\mathcal{G}^i}\gets\min\mathbb{E}_{\hat{x}^i\sim P(\hat{\textbf{X}}^i)}[1-{\rm log}D_i(\hat{x}^i)]$ 
		\Else
		\State Update $\Theta_{\mathcal{E}^i},\Theta_{\mathcal{G}^i}$ via the Eq.\ref{eq14}:
		\Statex \quad \quad \quad \quad \quad $\Theta_{\mathcal{E}^i}, \Theta_{\mathcal{G}^i}\gets\min\tilde{\mathcal{L}}_{ae}^{i}(\Theta_{\mathcal{E}^i};\Theta_{\mathcal{G}^i})$
		\State Update $\Theta_{\mathcal{D}^i}$ via the Eq.\ref{eq15}: 
		\Statex \quad \quad \quad \quad \quad $\Theta_{\mathcal{D}^i}\gets\max\mathbb{E}_{x^i\sim P(\textbf{X}^i)}[{\rm log}D_i(x^i)]$
		\Statex \quad \quad \quad \quad \quad \quad \quad \quad \quad \quad$+\mathbb{E}_{\hat{x}^i\sim P(\hat{\textbf{X}}^i)}[1-{\rm log}D_i(\hat{x}^i)]$
		\State Update $\Theta_{\mathcal{E}^i},\Theta_{\mathcal{G}^i}$ via the Eq.\ref{eq15}:
		\Statex \quad \quad \quad \quad \quad $\Theta_{\mathcal{G}^i}\gets\min\mathbb{E}_{\hat{x}^i\sim P(\hat{\textbf{X}}^i)}[1-{\rm log}D_i(\hat{x}^i)]$
		\State Calculate the common subspace \textbf{Z} by 
		\Statex \quad \quad \quad \quad \quad \quad \quad \quad $\textbf{Z}\gets\frac{1}{V}\sum_{i=1}^{V}{\textbf{Z}^i}$ 
		\State Update $\Theta_{\mathcal{D}^i}$ via the Eq.\ref{eq16}: 
		\Statex \quad \quad \quad \quad \quad $\Theta_{\mathcal{D}^i}\gets\max\mathbb{E}_{x^i\sim P(\textbf{X}^i)}[{\rm log}D_i(x^i)]$
		\Statex \quad \quad \quad \quad \quad \quad \quad \quad \quad \quad$+\mathbb{E}_{\tilde{x}^i \sim P(\tilde{\textbf{X}}^i)}[1-{\rm log}D_i(\tilde{x}^i)]$
		\State Update $\Theta_{\mathcal{E}^i},\Theta_{\mathcal{G}^i}$ via the Eq.\ref{eq16}:
		\Statex \quad \quad \quad \quad \quad $\Theta_{\mathcal{G}^i}\gets\min\mathbb{E}_{\tilde{x}^i \sim P(\tilde{\textbf{X}}^i)}[1-{\rm log}D_i(\tilde{x}^i)]$
		\EndIf
		\State Update the cluster centers $\textbf{\textit{U}}$ and cluster indicator
		\Statex \quad \ \ matrix $\textbf{\textit{S}}$ via the Eq.\ref{eq17}:
		\Statex \quad \quad \quad  \quad \quad $\textbf{\textit{U}},\textbf{\textit{S}}\gets \min\left \| \textbf{\textit{Z}}-\textbf{\textit{U}}\textbf{\textit{S}}\right \|_{F}^2$
		
		\EndFor
		\State \Return Multi-view encoder $\Theta_{\mathcal{E}^i}$ and the cluster $\textbf{\textit{U}}$, $\textbf{\textit{S}}$
	\end{algorithmic}
	\label{al2}
\end{algorithm}

Minimizing the auto-encoder loss to optimize our multi-view auto-encoder networks aims to learn the bidirectional mapping between the raw sample space and the common subspace. However, for each view, the $L_2$ distance function focuses on each sample dimension separately while ignoring the correlations between sample dimensions, which maybe lead to blurred reconstructed results and cannot model the sample distribution of each view. In the next, we detail the adversarial loss in our model to alleviate this problem.

\subsubsection{The Adversarial Loss}
Following GAN \cite{goodfellow2014generative}, for the $i$-th view, it consists of a generator $\mathcal{G}^i$ and a discriminator $\mathcal{D}^i$.  When $GS = 0$, for the $i$-th view, we suppose that the sample $x^i \sim P(\textbf{X}^i)$, and the generated one is $\hat{x}^i \sim P(\hat{\textbf{X}}^i)$. The adversarial loss is formulated as
\begin{equation}\label{eq15}
	\begin{aligned}
		\hat{\mathcal{L}}_{adv}^i(\Theta_{\mathcal{E}^i};\Theta_{\mathcal{G}^i};\Theta_{\mathcal{D}^i})&=\mathbb{E}_{x^i\sim P(\textbf{X}^i)}[{\rm log}D_i(x^i)]
		\\&+\mathbb{E}_{\hat{x}^i\sim P(\hat{\textbf{X}}^i)}[1-{\rm log}D_i(\hat{x}^i)],
	\end{aligned}
\end{equation}
where $D_i(\cdot)$ is required to distinguish $\textbf{X}^i$ and $\hat{\textbf{X}}^i$.
When $GS = 1$, the generated sample is $\tilde{x}^i \sim P(\tilde{\textbf{X}}^i)$ obtained by the common subspace $\textbf{Z}$, namely $\tilde{\textbf{X}}^i=G_i(\textbf{Z};\Theta_{\mathcal{G}^i})$. That can be described as
\begin{equation}\label{eq16}
	\begin{aligned}
		\tilde{\mathcal{L}}_{adv}^i(\Theta_{\mathcal{E}^i};\Theta_{\mathcal{G}^i};\Theta_{\mathcal{D}^i})&=\mathbb{E}_{x^i\sim P(\textbf{X}^i)}[{\rm log}D_i(x^i)]
		\\&+\mathbb{E}_{\tilde{x}^i \sim P(\tilde{\textbf{X}}^i)}[1-{\rm log}D_i(\tilde{x}^i)].
	\end{aligned}
\end{equation}
At the moment, in addition to the Eq. (\ref{eq15}),  $D_i(\cdot)$ is required to distinguish $\textbf{X}^i$ and $\tilde{\textbf{X}}^i$. By training the multi-view encoder networks and the
multi-view generator networks, we generate fake sample
similar to real sample of each view. The discriminators are
trained to distinguish the fake sample from the real sample
of each view. They play a min-max game until convergence.
We can obtain a common subspace similar to the
sample distribution of each view. According to the common subspace, we adopt a simple and effective k-means clustering algorithm to cluster different categories.

\subsubsection{The Clustering Loss}
Once multi-view common subspace $\textbf{\textit{Z}}=[z_1,...,z_n] \in \mathbb{R}^{p\times n}$ for $n$ data objects with $p$ dimensions is calculated, we obtain the multi-view clustering output by K-means algorithm via the following \cite{nie2019k,xia2020fast}:
\begin{equation}\label{eq17}
	\mathcal{L}_{clu} = \min_{\textbf{\textit{U}},\textbf{\textit{S}}}\left \| \textbf{\textit{Z}}-\textbf{\textit{U}}\textbf{\textit{S}}\right \|_{F}^2,
\end{equation}
\textit{s.t.}
\begin{center}
	$\textbf{\textit{S}} \in \{0,1\}^{c \times n}, \textbf{\textit{S}}^T\textbf{\textit{1}}=\textbf{\textit{1}},$
\end{center}
where $\textbf{\textit{U}}=[u_1,...,u_c] \in \mathbb{R}^{p\times c}$ denotes $c$ optimal cluster centers, $\textbf{\textit{\textit{S}}} \in \{0,1\}^{c \times n}$, where $\textbf{\textit{S}}_{q,k}$ indicates whether $z_k$ belongs to the $q$-th cluster. $\textbf{\textit{1}}$ is a vector with all entries as 1.  Algorithm \ref{al2} summarizes the above whole learning process for DAICS.

\section{Experiments}

\subsection{Experiment Setting}
\subsubsection{Datasets}
To demonstrate the performance of the proposed framework, we evaluate DAIMC and the compared baseline methods on four multi-view datasets. Tab. \ref{tab1} provides a brief description of each dataset. The details are described as follows. 

\begin{table}[b]
	\centering
	\caption{Dataset Descriptions.}  
	\label{tab1}    	
	\fontsize{8}{12}\selectfont  
	\begin{tabular}	{|m{2.5cm}<{\centering}|m{1.3cm}<{\centering}|m{1cm}<{\centering}|m{1cm}<{\centering}|}
		\hline
		Dataset & \#image   & \#view  & \#class \\ \hline
		\hline
		HW            & 2,000   & 2   & 10 \\  
		Caltech101-20 & 2,386   & 3   & 20 \\  
		NUS-WIDE-OBJ  & 3,100   & 2   & 31 \\  
		MNIST         & 70,000  & 2   & 10 \\ \hline
	\end{tabular}
\end{table}

\begin{itemize}
	\item \textbf{Handwritten numerals (HW)} \cite{asuncion2007uci}: This dataset  is composed of 2,000 data points from 0 to 9
		ten digit categories and each class has 200 data points. We adopt 76 Fourier coefficients of the character shapes and 216 profile correlations as two different views.	
	\item \textbf{Caltech101-20}: A subset of Caltech101 includes 2386 images of 20 subjects. We follow the setting used in \cite{zhao2017multi} to extract three handcrafted features including SIFT \cite{lowe2004distinctive} feature, HOG \cite{dalal2005histograms} feature and LBP \cite{ojala2002multiresolution} feature as three views. 		
	\item \textbf{NUS-WIDE-OBJ}: A subset of NUS-WIDE \cite{chua2009nus} consists of 30,000 images  distributed over 31 object categories. In the experiment, we randomly sample 100 images for each category and get 3100 images in total. We use two types of low-level features extracted from these images, including 64-D color histogram, 144-D color correlogram. 	
    \item \textbf{MNIST}: A widely-used large-scale benchmark data- set consisting of handwritten digit (0$\sim$9) images includes 70,000 samples with 28 $\times$ 28 pixels. We adopt the setting used in \cite{shang2017vigan}, the first view is the original images, and the other is given by images only highlighting the digit edge.
    %\vspace{-0.5em}
\end{itemize}

\renewcommand\arraystretch{1.5}
\begin{table*}[htp]	
	\centering
	\caption{Clustering performance of our proposed DAICS and baseline methods over HW, Caltech101-20 and NUS-WIDE-OBJ datasets. Results in bold face are the best for corresponding metrics.}
	\label{tab2}
	\fontsize{8}{10}\selectfont
	\begin{tabular}{m{2.25cm}<{\centering}|m{1.15cm}<{\centering}m{1.15cm}<{\centering}m{1.15cm}<{\centering}|m{1.15cm}<{\centering}m{1.15cm}<{\centering}m{1.15cm}<{\centering}|m{1.15cm}<{\centering}m{1.15cm}<{\centering}m{1.15cm}<{\centering}}
		\hline
		\multirow{2}{*}{Method} &
		\multicolumn{3}{c|}{HW}&\multicolumn{3}{c|}{Caltech101-20}&\multicolumn{3}{c}{NUS-WIDE-OBJ}\cr\cline{2-10}
		& ACC $\uparrow$    & NMI $\uparrow$    & Purity $\uparrow$  & ACC $\uparrow$    & NMI $\uparrow$    & Purity $\uparrow$ & ACC $\uparrow$    & NMI $\uparrow$    & Purity $\uparrow$  \cr
		\hline
		\hline
		SC$_{v=1}$ \cite{ng2002spectral}   &0.693    &0.674    &0.708    &0.319    &0.447    &0.317    &0.153    &0.155    &0.152     \cr
		SC$_{v=2}$ \cite{ng2002spectral}     &0.658    &0.651    &0.672    &0.313    &0.416    &0.310    &0.149    &0.141    &0.145     \cr
		ClusterGAN$_{1}$ \cite{mukherjee2019clustergan} &0.842    &0.829    &0.855    &0.346    &0.445    &0.342    &0.191    &0.184    &0.187     \cr
		ClusterGAN$_{2}$  \cite{ghasedi2019balanced} &0.856    &0.833    &0.869    &0.378    &0.496    &0.376    &0.183    &0.188    &0.179     \cr
		DCCA  \cite{andrew2013deep}  &0.860    &0.842    &0.877    &0.428    &0.620    &0.426    &0.205    &0.196    &0.203     \cr
		MvDMF \cite{zhao2017multi}  &0.843    &0.827    &0.854    &0.362    &0.487    &0.360    &0.185    &0.177    &0.182     \cr
		DAMC  \cite{li2019deep}      &0.954    &0.920    &0.961    &0.534    &0.661    &0.531    &0.243    &0.235    &0.241     \cr
		EAMC  \cite{zhou2020end}     &0.946    &0.925    &0.958    &0.557    &0.674    &0.553    &0.255    &0.238    &0.252     \cr
		MvSCN \cite{huang2019multi}  &0.965    &\bf0.957 &0.972    &0.578    &0.690    &0.574    &0.226    &0.221    &0.223     \cr
		\hline
		\hline
		DAICS (ours)         &\bf0.974 &0.951    &\bf0.983 &\bf0.605 &\bf0.728 &\bf0.603 &\bf0.278 &\bf0.265 &\bf0.263  \cr
		\hline
	\end{tabular}
\end{table*}

\begin{figure*}[htp]
	\quad \ 
	\begin{subfigure}{.3\textwidth}
		\centering
		\includegraphics[width=\textwidth]{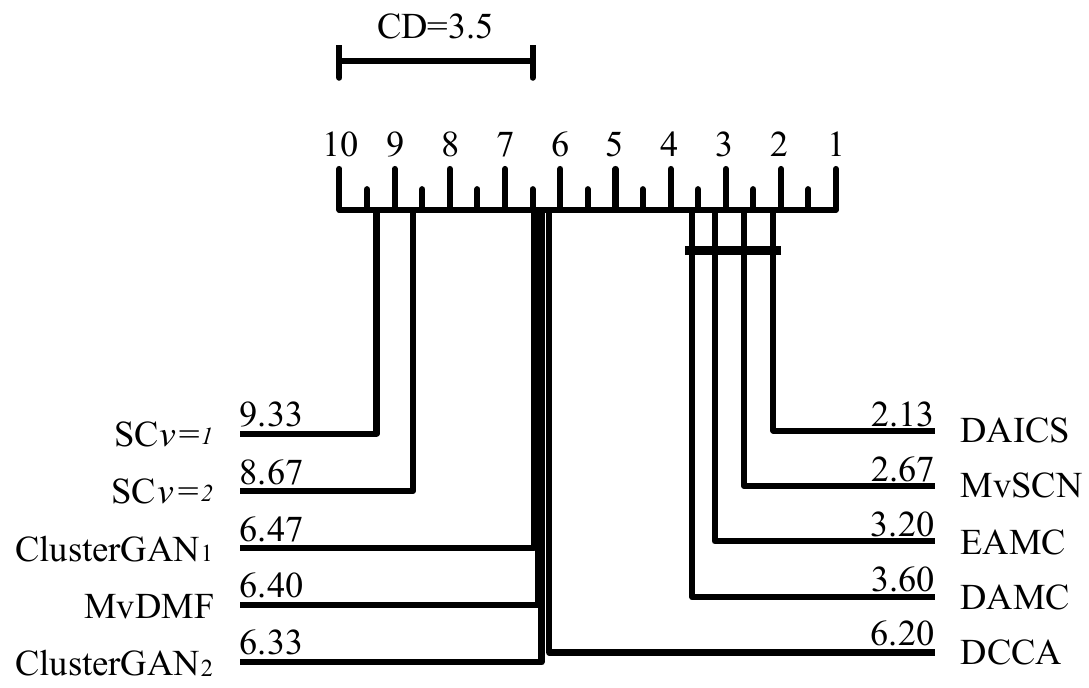}
		\caption{}
	\end{subfigure}
	\quad
	\begin{subfigure}{.32\textwidth}
		\centering
		\includegraphics[width=\textwidth]{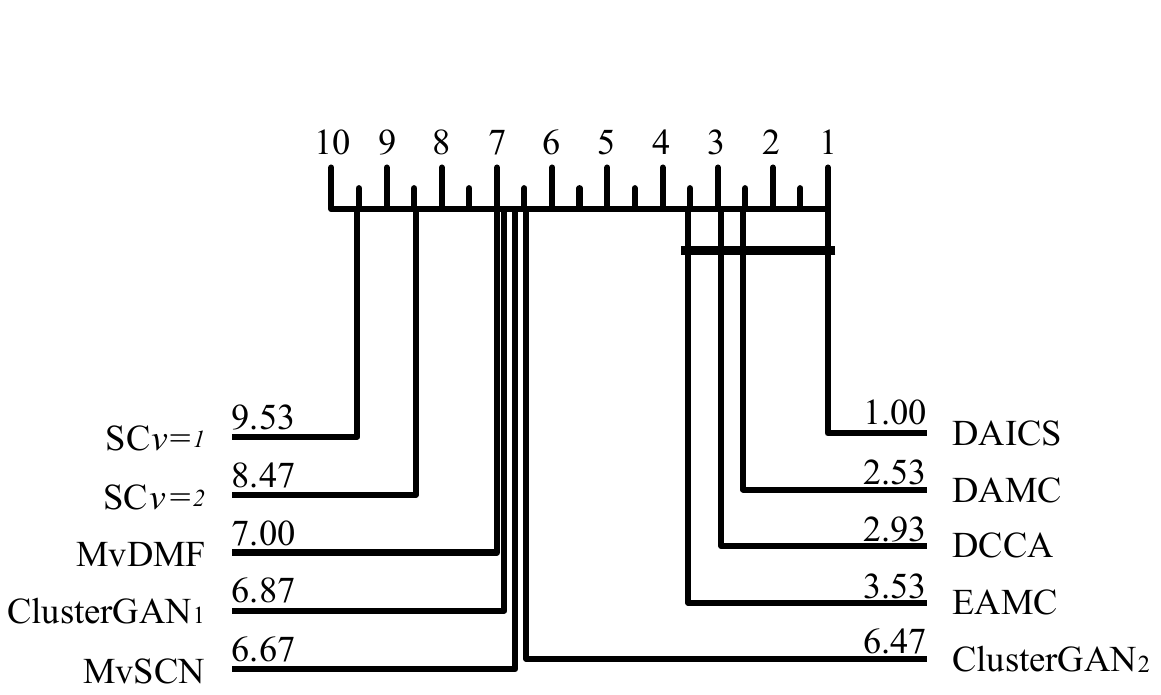}
		\caption{}
	\end{subfigure}
	\quad
	\begin{subfigure}{.3\textwidth}
		\centering
		\includegraphics[width=\textwidth]{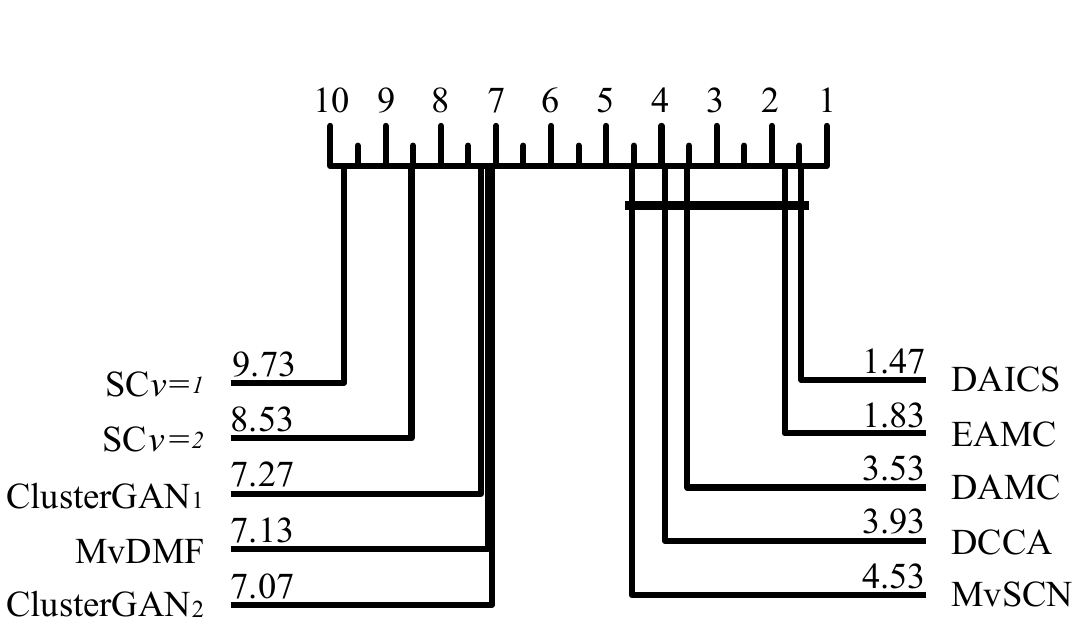}
		\caption{}
	\end{subfigure}
	\caption{Comparisons of our proposed DAICS against baseline methods using the Nemenyi test on 15 folds of per dataset ((a) HW, (b) Caltech101-20, (c) NUS-WIDE-OBJ). The methods that are not significantly different at 0.05 level are connected.}
	\label{fig4}
	%\vspace{-1em}	
\end{figure*}

\subsubsection{Evaluation Metrics}
The clustering performance is measured by using three standard evaluation matrices, \emph{i.e.,} Accuracy (\textbf{ACC}), Normalized Mutual Information (\textbf{NMI}), and \textbf{Purity}. For all metrics, the higher value indicates the better performance. More details could be found in \cite{kumar2011co}.
\subsubsection{Comparison methods}
To exhibit the superiority of DAICS, we adopt spectral clustering \cite{ng2002spectral} (following \cite{li2019deep,huang2019multi}, we test the spectral clustering with 1 and 2 views, denoted as SC$_{v = 1}$ and SC$_{v = 2}$) and the state-of-the-art deep clustering algorithms as baseline models, including
recent deep multi-view clustering methods: Deep Canonical Correlation Analysis (DCCA) \cite{andrew2013deep}, Multi-view clustering via Deep Matrix Factorization (MvDMF) \cite{zhao2017multi}, Deep Adversarial Multi-view Clustering network (DAMC) \cite{li2019deep}, End-to-End Adversarial-Attention network for Multi-Modal Clustering (EAMC) \cite{zhou2020end}, Multi-view Spectral Clustering Network (MvSCN) \cite{huang2019multi}. To validate our merits, we also test typical single-view deep clustering methods: latent space Clustering in Generative Adversarial Networks (ClusterGAN$_{1}$) \cite{mukherjee2019clustergan}, Generative Adversarial Clustering Network (ClusterGAN$_{2}$) \cite{ghasedi2019balanced}.
\subsubsection{Implementation Details} All the experiments are implemented by using the public toolbox of PyTorch on a standard Ubuntu-16.04 OS with NVIDIA 1080Ti GPUs and 64 GB memory size. We adopt the \textbf{Adam} optimizer as our optimization method with its parameters $\beta_1 = 0.5$, $\beta_2 = 0.99$ and learning rate to be 1e-4. Instead of manually setting $\lambda(t)$, we choose it based on the loss values of samples such that we select only $5\%$ of all samples as input samples to initialize training process, and then decrease $\lambda(t)$ to include all samples at 4/5 of the maximum epoch.

\subsection{Experimental Results}
\subsubsection{Compared with State of the Arts}
We compare DAICS with the baseline methods. Tab. \ref{tab2} shows the clustering results of DAICS and other methods on three datasets. It is obvious that the clustering results of deep multi-view clustering methods significantly outperform the single-view clustering methods. Specifically, our DAICS achieves 97.4\% accuracy on HW dataset, which is the best performance. In addition, DAICS also outperforms other deep methods with a clear improvement on both Caltech101-20 and NUS-WIDE-OBJ datasets. Fig. \ref{fig4} shows the results from the Nemenyi test described when the analysis is performed per dataset. Overall, it is observed that the average rank of our proposed DAICS method is higher than those of the others, and the results perform statistically significant. We attribute this success to adversarial inconsistent samples, cognitive sampling and golden section among multiple views.  
\begin{table}[htb]
	\centering
	\caption{Clustering results on large-scale MNIST dataset.  Results in bold face are the best for corresponding metrics.}
	\label{tab3}
	\fontsize{8}{10}\selectfont
	\begin{tabular}	{m{1.8cm}<{\centering}|m{1.2cm}<{\centering}m{1.2cm}<{\centering}m{1.2cm}<{\centering}}
		\hline
		Method   & ACC $\uparrow$      & NMI $\uparrow$       & Purity $\uparrow$     \\ \hline
		\hline
		DCCA  \cite{andrew2013deep}    & 0.472    & 0.437    & 0.496     \\ \hline
		DAMC  \cite{li2019deep}    & 0.647    & 0.598    & 0.653     \\ \hline
		EAMC  \cite{zhou2020end}    & 0.665    & 0.614    & 0.649     \\ \hline
		DAICS (ours)    & \bf0.698 & \bf0.651 & \bf0.693  \\ \hline
	\end{tabular}	
\end{table}

\subsubsection{Clustering on Large-scale Dataset}
To test the efficiency of DAICS on large-scale dataset, we compare DAICS with three deep models (DCCA, DAMC and EAMC) on MNIST dataset. It is worth noting that, DAICS is efficient due to the cognitive sampling strategy for multi-view clustering. As shown in Tab. \ref{tab3}, our proposed method consistently outperforms the other methods in terms of ACC, MNI and Purity, which validates the superiority of our DAICS on the large-scale dataset.\begin{figure}[htb]
	\includegraphics[width=0.7\columnwidth]{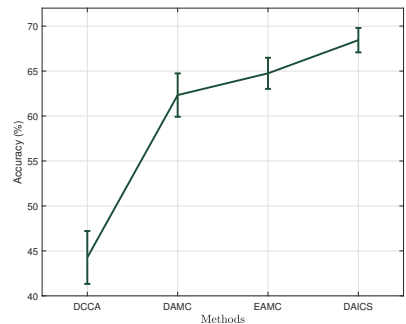}
	\centering
	\caption{The mean and standard deviation of the accuracy of DAICS and baseline methods on MNIST dataset.}
	\label{fig5}
\end{figure}  

\begin{figure*}[htp]
	\quad \ 
	\begin{subfigure}{.13\textwidth}
		\centering
		\includegraphics[width=\textwidth]{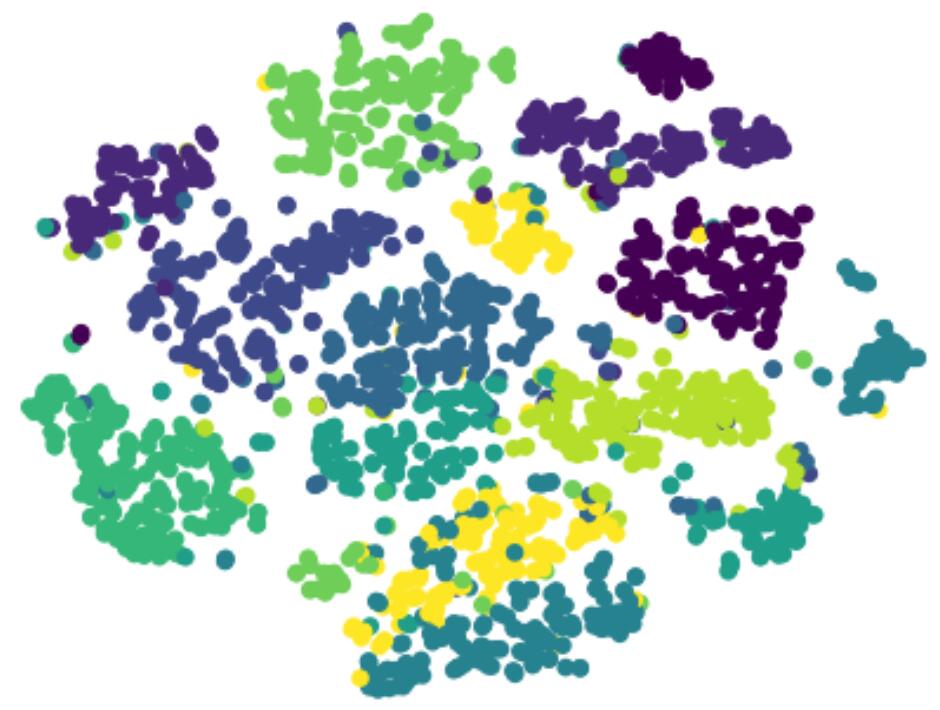}
		\caption{View 1}
	\end{subfigure}
	\quad
	\begin{subfigure}{.13\textwidth}
		\centering
		\includegraphics[width=\textwidth]{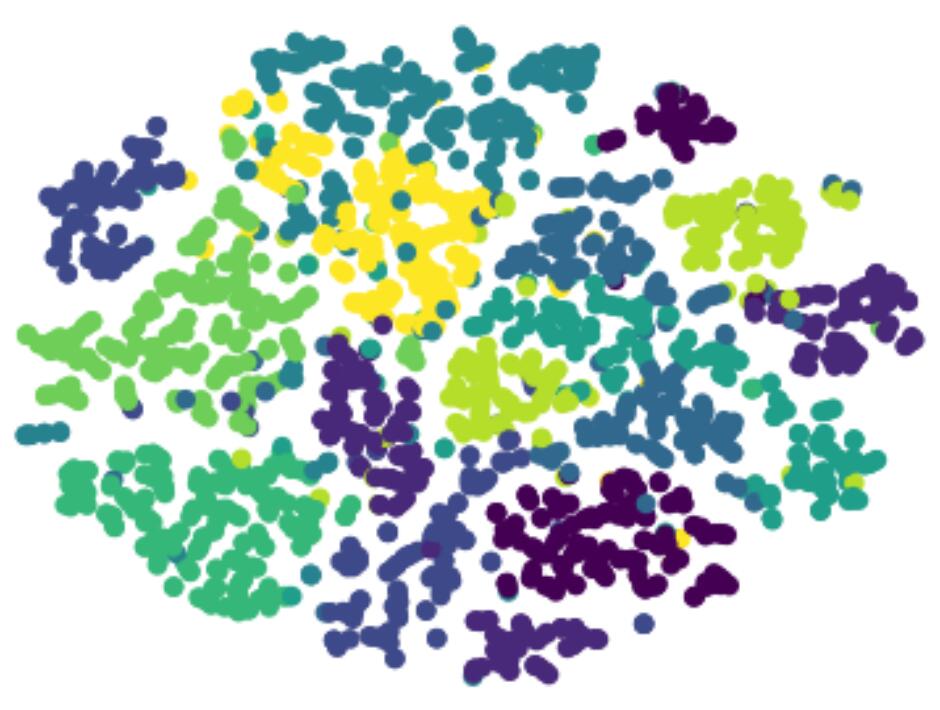}
		\caption{View 2}
	\end{subfigure}
	\quad
	\begin{subfigure}{.14\textwidth}
		\centering
		\includegraphics[width=\textwidth]{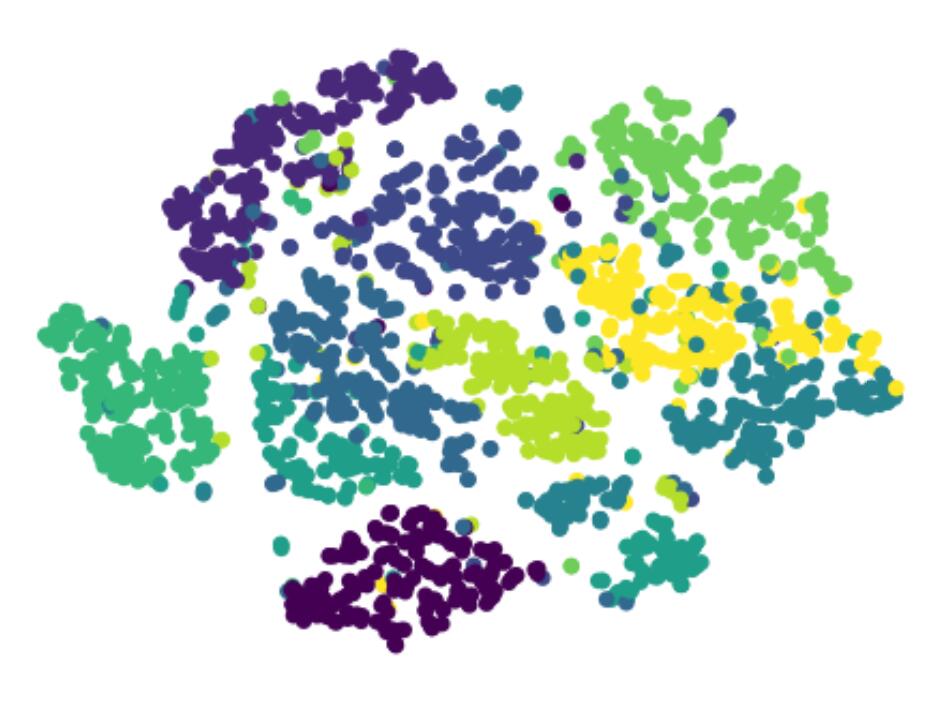}
		\caption{DCCA}
	\end{subfigure}
	\quad
	\begin{subfigure}{.14\textwidth}
		\centering
		\includegraphics[width=\textwidth]{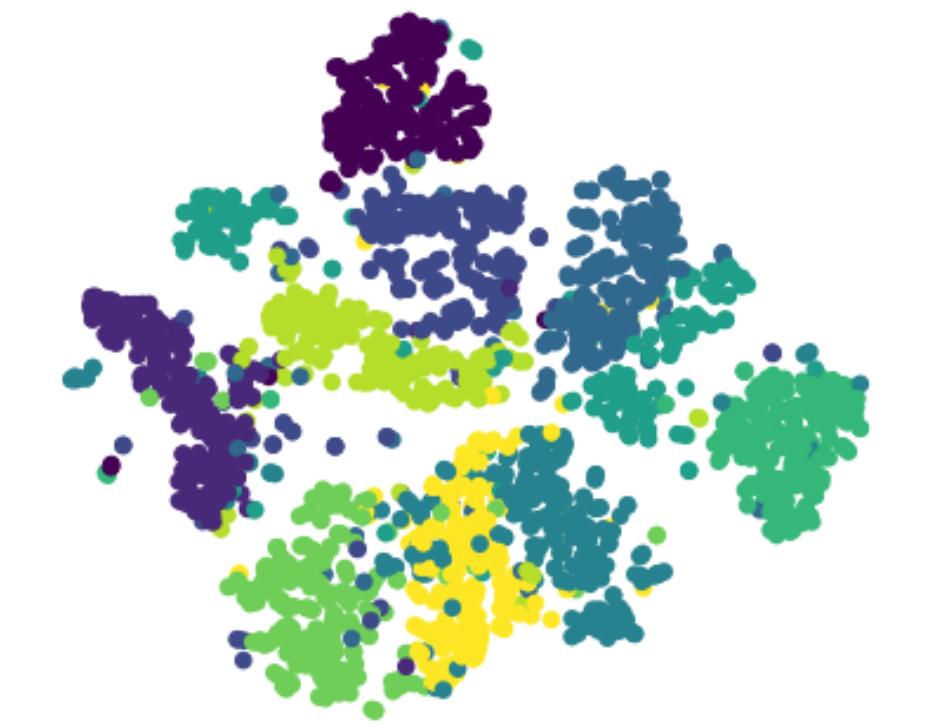}
		\caption{DAMC}
	\end{subfigure}
	\quad
	\begin{subfigure}{.14\textwidth}
		\centering
		\includegraphics[width=\textwidth]{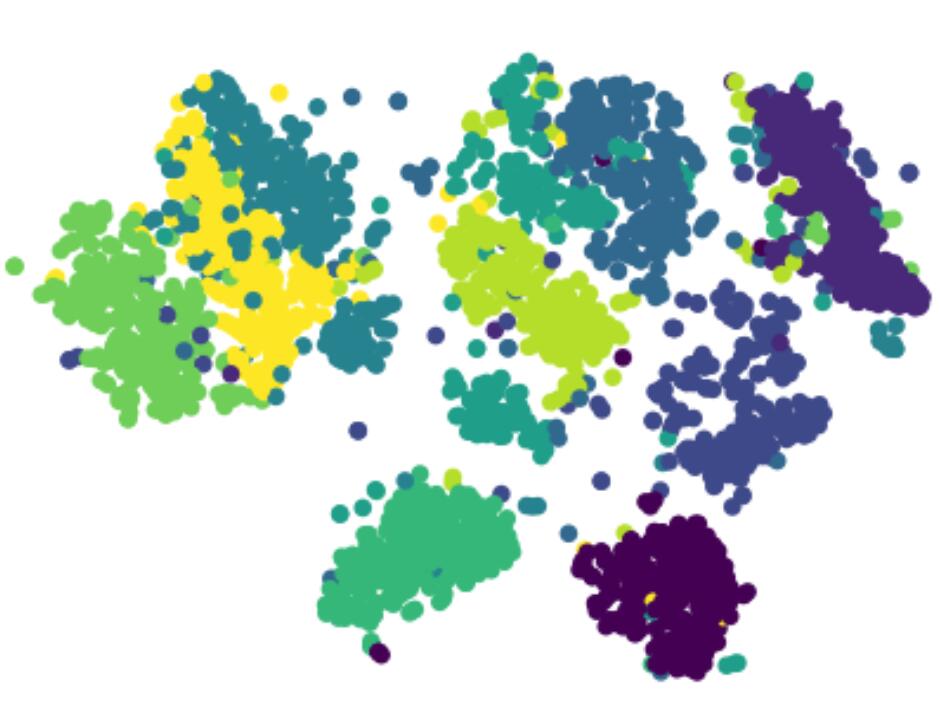}
		\caption{EAMC}
	\end{subfigure}
	\quad
	\begin{subfigure}{.14\textwidth}
		\centering
		\includegraphics[width=\textwidth]{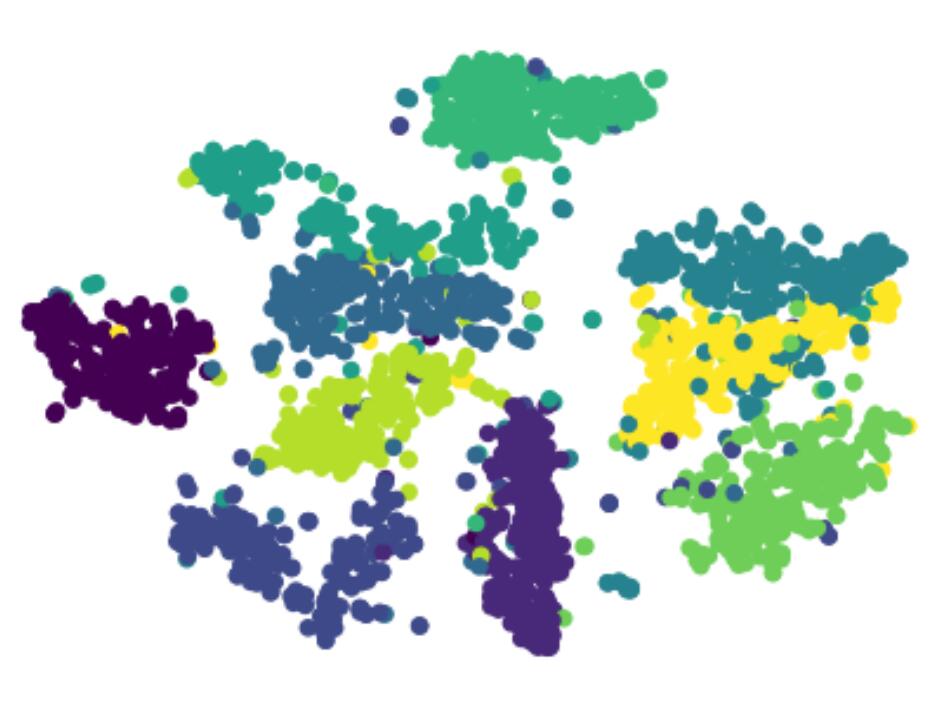}
		\caption{DAICS}
	\end{subfigure}
	\caption{Visualization of deep multi-view clustering results of four methods ((c) DCCA; (d) DAMC; (e) EAMC; (f) Our proposed DAICS.) via the t-SNE on MNIST dataset with two views: (a) Original features as the first view; (b) Edge features as the second view. We observe that DAICS offers a more clear and compact cluster structure than other methods.}
	\label{fig6}
	%\vspace{-1em}	
\end{figure*} 
In order to compare the results more fairly, we repeated the experiment of our proposed DAICS and baseline methods on MNIST dataset for 15 times. The mean and standard deviation of the accuracy of them are shown in Fig. \ref{fig5} and our proposed DAICS outperforms obviously the baseline methods. To further demonstrate the above fact, we present the visualized results of all these methods on 2500 samples randomly selected from MNIST via the t-SNE. As it is widely known that the visualized results of the t-SNE are more uniform than the UMAP. Therefore, we only adopt the t-SNE to achieve the visualized results as shown in Fig. \ref{fig6}. It can be easily seen that DAICS offers a more clear and compact cluster structure than other deep models.

\subsubsection{Ablation Study}
To further validate the effectiveness of each component for DAICS, extensive ablation studies are performed,  including the adversarial inconsistent samples (AIS), the cognitive sampling (CS), and the golden section (GS). For brevity, DAICS$_{NONE}$ denotes DAICS without the above three components. DAICS$_{CS}$ denotes DAICS with the CS only, since there is no the AIS, we adopt the input samples (from easy to difficult) of the training network with the view that achieves the best performance. DAICS$_{CS+GS}$ makes up the GS to DAICS$_{CS}$. DAICS$_{AIS+CS}$ denotes DAICS without the GS. Tab. \ref{tab4}.  Similarly, we repeated the experiment of our proposed DAICS and different variants on HW dataset for 15 times. Fig. \ref{fig7} show the mean and standard deviation of the results of the above models on HW dataset. All of the components are effective, due to the following observations:
\begin{figure}[b]
	\includegraphics[width=0.7\columnwidth]{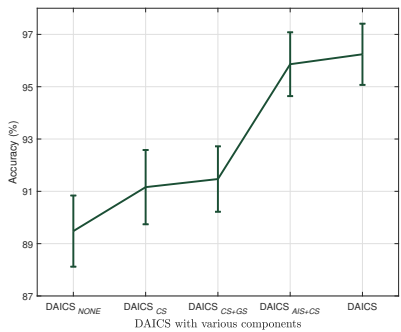}
	\centering
	\caption{The mean and standard deviation of the accuracy of DAICS with various components on HW dataset.}
	\label{fig7}
\end{figure} 

\begin{table}[htb]
	\centering
	\caption{Ablation study on HW dataset. Results in bold face are the best for corresponding metrics.}
	\label{tab4}
	\fontsize{8}{10}\selectfont
	\begin{tabular}	{m{2.35cm}|m{1.2cm}<{\centering}m{1.2cm}<{\centering}m{1.2cm}<{\centering}}
		\hline
		Model            & ACC $\uparrow$       & NMI $\uparrow$      & Purity $\uparrow$     \\ \hline
		\hline
		DAICS$_{NONE}$   &0.908      &0.892      & 0.915       \\ \hline
		DAICS$_{CS}$     &0.926      &0.915      & 0.934       \\ \hline
		DAICS$_{CS+GS}$  &0.927      &0.912      & 0.937       \\ \hline
		DAICS$_{AIS+CS}$ &0.971      &0.946      & 0.976       \\ \hline
		DAICS            &\bf0.974   &\bf0.951   &\bf0.983     \\ \hline
	\end{tabular}
	\vspace{-1em}
\end{table}

\begin{figure}[htb]
	\includegraphics[width=0.65\columnwidth]{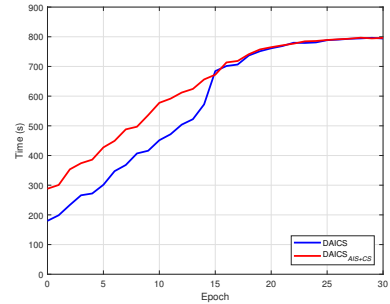}
	\centering
	\caption{The effect of the GS on efficiency of DAICS over varied training epochs on HW dataset.}
	\label{fig8}
	\vspace{-0.5em}
\end{figure}

\begin{itemize}
	\item Upon the results of DAICS$_{NONE}$, DAICS$_{CS}$ and DAICS$_{AIS+CS}$, it can be easily seen that the AIS and CS are key components that can avoid getting stuck in non-ideal local minima for the better clustering results.

	\item Despite the improvement of the GS on clustering results is not obvious from DAICS$_{AIS+CS}$ and DAICS, it can be observed that the GS significantly improves the network efficiency as per Fig. \ref{fig8}.
\end{itemize}

\subsubsection{Parameters Analysis}
\begin{figure}[htb]
	\centering
	\subcaptionbox{}{
		\centering
		\includegraphics[width=1.4in]{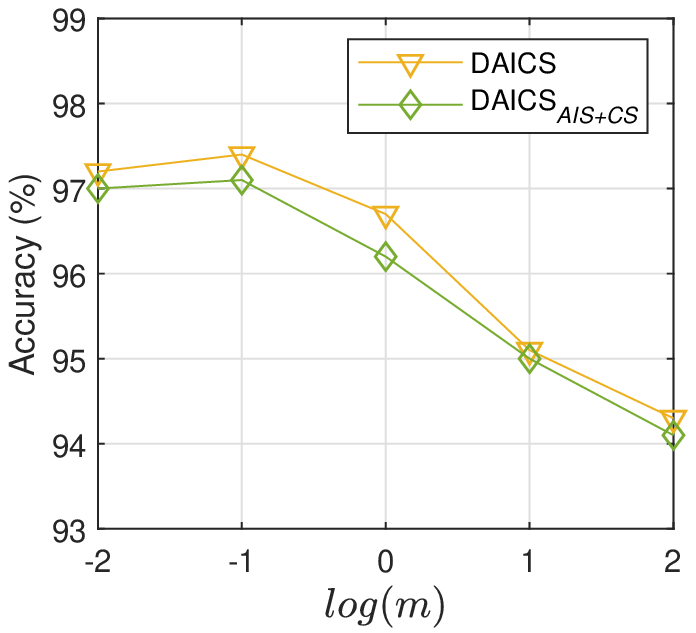}
	}%
	\subcaptionbox{}{
		\centering
		\includegraphics[width=1.4in]{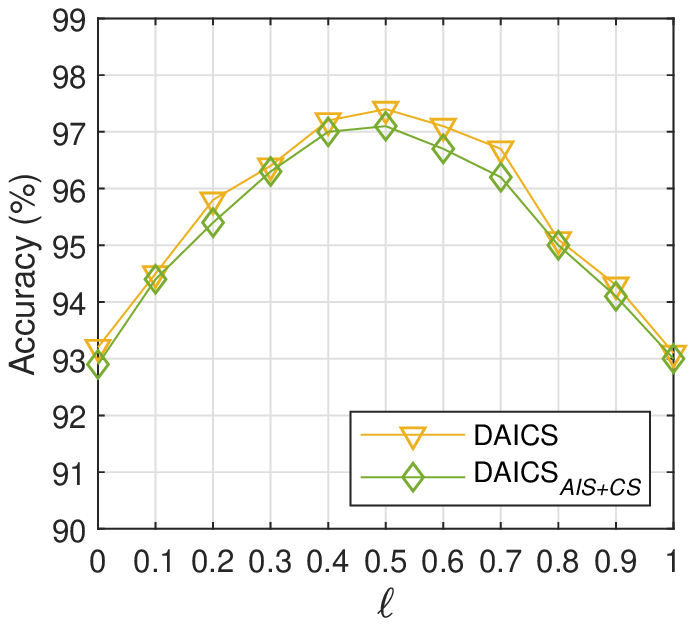}
	}%
	\caption{Clustering performance of DAICS$_{AIS+CS}$ and DAICS with varied values for parameters: (a) The margin $m$; (b) The factor $\ell$ on HW dataset.}
	\label{fig9}
	\vspace{-0.5em}
\end{figure}

We conduct parameter analysis of DAICS on HW dataset, including $m, \ell, \alpha$ and $\beta$ for our essential AIS module on multi-view clustering. Specifically, the feature similarity loss $\mathcal{L}_{sim}$ and the  binary classification loss $\mathcal{L}_{adv}$ are influenced by the margin $m$ and the factor $\ell$, respectively. As shown in Fig. \ref{fig9}(a), when $m \in [0.01,0.1]$, our DAICS can obtain the better clustering performance. Meanwhile, $\ell = 0.5$  is more suitable for DAICS as demonstrated in Fig. \ref{fig9}(b).
 \begin{figure}[htb]
	\centering
	\subcaptionbox{}{
		\includegraphics[width=1.5in]{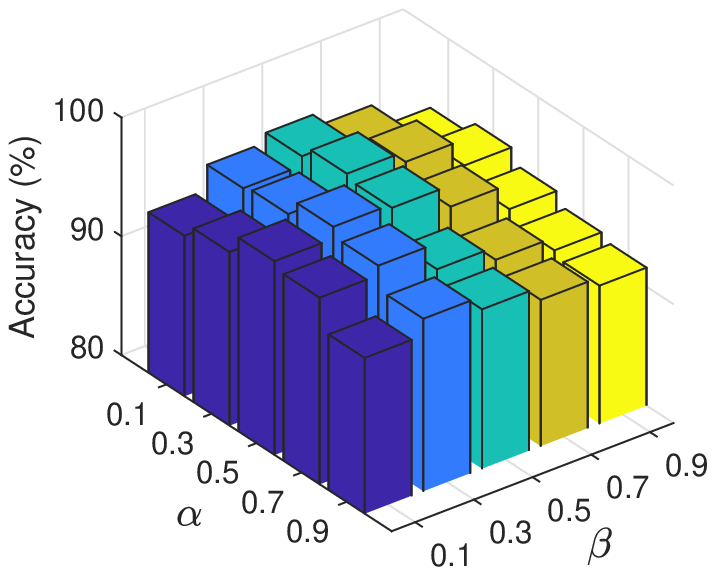}
	}%
	\subcaptionbox{}{
		\includegraphics[width=1.6in]{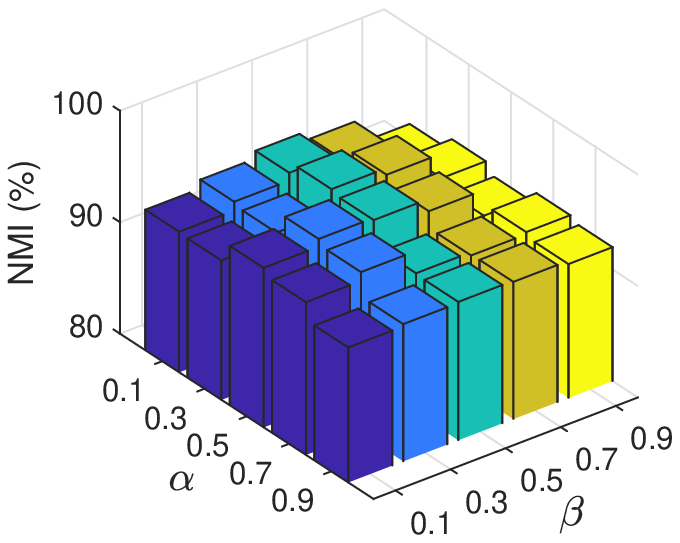}
	}%
	\caption{The effect of the parameters $\alpha$ and $\beta$ on clustering performance against HW dataset. (a) and (b) are the clustering results in terms of ACC and NMI.}
	\label{fig10}
	\vspace{-0.5em}
\end{figure}
Besides, the adjustable factors $\alpha$ and $\beta$ can control the optimization process with an adversarial minimax game of the AIS module. When $\alpha = 0.3$ and $\beta = 0.5$, it can be seen that the best clustering result is obtained about ACC and NMI as shown in Fig. \ref{fig10}(a) and Fig. \ref{fig10}(b).

\section{Conclusion}
In this paper, we propose a novel Deep Adversarial Inconsistent Cognitive Sampling (DAICS) method for multi-view progressive subspace clustering. DAICS consists of the AIS module, the CS strategy and multi-view progressive clustering network with the GS mechanism. The AIS module exploits an adversarial minimax game of the binary classification loss and the feature similarity loss for sample consistency. The CS strategy gradually selects the input samples from easy to difficult for multi-view clustering network training. Moreover, the GS mechanism is developed for efficiency. Experimental results demonstrate that DAICS outperforms the state-of-the-arts over real-world datasets.

\ifCLASSOPTIONcaptionsoff
  \newpage
\fi

%
%\end{thebibliography}
\bibliographystyle{ieeetr}
\bibliography{ref}


%\newpage


%\vfill


\end{document}